%% file: main.tex
\documentclass[12pt,arial,onecolumn,letterpaper]{elsarticle}
\usepackage[margin=0.8in]{geometry}
\usepackage{url}
\usepackage{xcolor}
\usepackage{amsmath}
\usepackage{amssymb}
\usepackage{xspace}
\usepackage{booktabs}
\usepackage{enumerate}
\usepackage{mathtools}
\usepackage{algorithm,algorithmic}
\usepackage{graphicx}
\usepackage{subcaption}

\definecolor{newcolor}{rgb}{.8,.349,.1}
\newcommand{\ie}{i.e.~}

\newcommand{\etal}{et al.~}
\newtheorem{theorem}{Theorem}[section]
\newtheorem{lemma}[theorem]{Lemma}
\newtheorem{definition}[theorem]{Definition}

\newtheorem{corollary}[theorem]{Corollary}

\graphicspath{{./}{./}}

\def\Vec#1{{\boldsymbol{#1}}}
\def\Mat#1{{\boldsymbol{#1}}}

\makeatletter
\newcommand{\ALOOP}[1]{\ALC@it\algorithmicloop\ #1%
  \begin{ALC@loop}}
\newcommand{\ENDALOOP}{\end{ALC@loop}\ALC@it\algorithmicendloop}

\makeatother

\def \rev#1{{#1}}

\usepackage[pagebackref=false,colorlinks,bookmarks=false,urlcolor=blue,linkcolor=blue,citecolor=blue,pdftitle={Efficient Clustering on Riemannian Manifolds: A Kernelised Random Projection Approach},pdfauthor={Kun Zhao, Azadeh Alavi, Arnold Wiliem, Brian C. Lovell}]{hyperref}

\begin{document}

\begin{frontmatter}

\title{\textbf{Efficient Clustering on Riemannian Manifolds:\\ A Kernelised Random Projection Approach}}

\author{
		Kun Zhao{\tiny~}$^{\star}$, Azadeh Alavi{\tiny~}$^{\dagger}$, Arnold Wiliem{\tiny~}$^{\star}$, Brian C. Lovell{\tiny~}$^{\star}$ \\
$^{\star}$School of ITEE, The University of Queensland, St Lucia, Brisbane, QLD 4072, Australia\\		
$^{\dagger}$ Center for Automation Research, University of Maryland, College Park, MD 20742-3275 \\
{\tt\small kun.zhao@uq.net.au, azadeh@umiacs.umd.edu, a.wiliem@uq.edu.au, lovell@itee.uq.edu.au}
}




 
\begin{abstract}
Reformulating computer vision problems over Riemannian manifolds has demonstrated superior performance in various computer vision applications.
This is because visual data often forms a special structure lying on a lower dimensional space embedded in a higher dimensional space.
However, since these manifolds belong to non-Euclidean topological spaces, exploiting their structures is computationally expensive, especially when one considers the clustering analysis of massive amounts of data.
To this end, we propose an efficient framework to address the clustering problem on Riemannian manifolds.
This framework implements random projections for manifold points via kernel space, which can preserve the geometric structure of the original space, but is computationally efficient. 
Here, we introduce three methods that follow our framework.
We then validate our framework on several computer vision applications by comparing against popular clustering methods on Riemannian manifolds.
Experimental results demonstrate that our framework maintains the performance of the clustering whilst massively reducing computational complexity by over two orders of magnitude in some cases.
\end{abstract}


\end{frontmatter}
\input{introduction}
\input{Riemannian-Manifolds}
\input{methodology}

\input{results}

\input{conclusion}
\input{Acknowledgements}
{\small
\bibliography{refs}
}

\input{cv}

\end{document}

%% file: introduction.tex
\section{Introduction}
\label{sec1}
Clustering analysis is an automated process that groups unlabelled data into subsets (here called clusters) that may express the underlying structure of the data.
It is one of the most critical tools for understanding visual data~\cite{jain2010,dhillon2004kernel}.
For instance, significant amounts of visual data such as videos and pictures are uploaded every second~\cite{Doe2012Online}.
Indeed, this is the case for YouTube where 100 hours of video are uploaded every minute~\cite{Doe2014youtube}.
Although these videos have titles and some additional meta-information, it is often desirable to automatically group the videos in terms of 
specific criteria such as visual similarity or detected objects.

In recent years, modelling visual data in analytical manifolds such as Riemannian manifolds has enjoyed success in 
various computer vision application domains such as face recognition~\cite{Turagaetal2011},
action recognition~\cite{harandi2013kernel} and pedestrian detection~\cite{tuzel2008pedestrian}.
This is because visual features and models often possess special structures which Euclidean space fails to capture.
Riemannian manifolds which form curved spaces are a more appropriate approach to model problems in various computer vision tasks.  

Unfortunately, despite the fact that clustering methods have been studied since the $1950s$~\cite{jain2010,filippone2008survey}, 
applying such methods directly on data represented on Riemannian manifolds is not trivial.
Riemannian manifolds generally do not conform to Euclidean space~\cite{Turagaetal2011,pennecetal2006}. To address this,  
one could use manifold tangent spaces which are locally homeomorphic to Euclidean space~\cite{pennecetal2006}.
However, this brings another challenge to applying existing clustering algorithms as some general algebraic operations 
are not well defined~\cite{pennec2006intrinsic}.
For instance, K-means requires the computation of the mean within a cluster which cannot be computed directly.
To this end, Pennec~\etal\cite{pennec2006intrinsic} reformulated the computation of mean as a solution to an optimisation problem.
Using this formulation, the mean point is considered as the point over the manifold minimising the geodesic distance (\ie the true distance on the manifold between two points) from the mean point to all other points.
The algorithm to solve this problem is called Karcher mean~\cite{pennec2006intrinsic}.
Thanks to the Karcher mean, Turaga~\etal\cite{Turagaetal2011} extended the K-means algorithm into the Riemannian manifold, which is regarded as intrinsic K-means and has been applied to activity-based video clustering.
Intrinsic K-means has further demonstrated better performance than Euclidean-based methods (for example, 
Protein Clustering~\cite{suryanto2012protein}).

Generally, methods that completely honour the manifold topology lead to higher accuracy.
We shall categorise these methods as intrinsic methods.
Unfortunately, the computational cost of intrinsic methods is extremely high since these need to map all of the data to tangent spaces repeatedly.

Extrinsic methods, on the other hand, seek solutions that may not completely consider the manifold topology~\cite{faraki2014log,faraki2014fisher,yuanetal2010,jayasumanaetal2013,jayasumana2013framework,harandi2014expanding}.
The most simplistic way, here called Log Euclidean methods, is to embed all of the points into a designated tangent space at the identity point~\cite{arsignyetal2006}.
Log Euclidean methods can be considered as flattening the manifold space. It has been used in various computer vision applications, such as human action recognition~\cite{faraki2014fisher} and cell classification~\cite{yuanetal2010}.
This addresses the computational cost issues suffered by the intrinsic methods, as the tangent space is homeomorphic to the Euclidean space and well-known Euclidean clustering approaches such as K-means can be directly applied.
Unfortunately, as the flattening step distorts the pair-wise distances in regions far from the origin of the tangent space, accuracy is severely compromised.
So much of the value of the manifold approach is lost.

Other approaches that fall in the extrinsic method category are kernel-based approaches~\cite{jayasumanaetal2013,jayasumana2013framework,harandi2014expanding}, such as Kernel K-means.
In essence, the data in manifold space are first embedded into the Reproducing Kernel Hilbert Space (RKHS)~\cite{shaweCristianini2004}. 
As the embedding function is defined implicitly, generally kernel-based approaches make use of the inner products in the RKHS in their formulation.
These inner products are then arranged in a Gram matrix.
It is often observed that the right choice of kernel could significantly improve the performance~\cite{jayasumanaetal2013}.
Furthermore, in general, kernel inner products with specified metrics have much less computational complexity than geodesic distances~\cite{alavietal2014,shirazietal2012}.
With these properties, kernel-based approaches could be suitable to address issues suffered in both the intrinsic approach and the Log Euclidean approach.
Unfortunately, the kernel-based approaches cannot scale easily, 
as the Gram matrix computation is $O(n^2)$ where $n$ is the number of data points. 
Also, it is often quite challenging to kernelise the existing algorithms that do not have known kernelised versions~\cite{caseiro2013rolling}.
Furthermore, Nikhil~\etal demonstrate that clustering data in the RKHS may lead to unexpected results since the clusters obtained in the RKHS may not exhibit the structure of the original data\cite{pal2014and}.

\textbf{Contributions}
We summarise the advantages and shortcomings of the existing approaches in Table~\ref{tab:related work}.
Our goal is to develop an efficient clustering algorithm for Riemannian manifolds, which significantly reduces the computational complexity, but still maintains acceptable performance.
The inspirations are drawn from the random projection for Euclidean spaces which has enjoyed success in various domains~\cite{bingham2001random,kushilevitz2000efficient,goel2005face}
due to its simplicity and theoretical guarantees~\cite{achlioptas2003database}. 
We list our contributions as follows:
\begin{enumerate}
\item We propose a random projection framework for manifold features.
In general, the term projection is not well defined in Riemannian manifolds.
Therefore, we address this via 
the RKHS constructed from a small subset of data.
Once projected, we choose to apply the 
K-means algorithm.
\item From our framework, it becomes clear that
random hyperplane generation is essential. Thus, we describe three generation algorithms which are followed in our framework:
(1) Kernelised Gaussian Random Projection (KGRP); 
(2) Kernelised Orthonormal Random Projection (KORP) 
and (3) Kernel Principal Component Analysis 
Random Projection (KPCA-RP).
\end{enumerate}  


We note that our method is different from manifold learning approaches for clustering analysis described in~\cite{elhamifar2011sparse}.
Manifold learning is the collection of 
non-linear dimensionality reduction (NLDR) techniques that seek for a low dimensional
representation of a set of high-dimensional points lying on a non-linear manifold~\cite{lin2006riemannian}.
They assume the structure of the underlying manifold was unknown.
Contrary to this, in our paper, we
are interested in Riemannian manifolds whose underlying geometry is known.



\begin{table}
    \centering
      \caption
      {Summary of the existing works compared to our proposal.}
      \label{tab:related work}
       \vspace{0.5ex}    
       {
          \begin{tabular}{cccp{2cm}}
                 \toprule
                 ~{Approach}~&
                 { Exploits Manifold Structure}&
                 {Accuracy}~& 
                 {Computational} \\
                 {}&&&{Complexity}\\
                  \toprule
                 {Intrinsic Methods\cite{Turagaetal2011,suryanto2012protein}}
                 & {Yes} 
                 &{ High} 
                 &{ High}
                 \\
                 {Log-Euclidean Methods~\cite{faraki2014log,faraki2014fisher,yuanetal2010}}
                 &{Minimal}
                 &{Low}
                 &{Low}
                 \\
                 {Kernel Methods~\cite{jayasumanaetal2013,jayasumana2013framework,harandi2014expanding}}
                 &{Approximately}
                 &{High}
                 &{Moderate}\\
                 {Our proposal}
                 &{Approximately}
                &{High}
                 &{Low}\\
                 \bottomrule
                  \end{tabular}
                }  
    \end{table}

We continue the paper as follows.
Section~\ref{sec:riemannian_geometry} provides a brief mathematical background of Riemannian manifolds.
Section~\ref{sec:proposed_framework} details the proposed random projection framework for manifold points and develops three different random projection methods for clustering points on manifold spaces.  
The proposed methods are then contrasted with the state-of-the-art methods in Section~\ref{sec:experimental_results}.
The conclusions and future directions are summarised in Section~\ref{sec:conclusions}. 

%% file: Riemannian-Manifolds.tex
\section{The Geometry of Riemannian Manifolds}
\label{sec:riemannian_geometry}
A differentiable manifold $\Mat{\mathcal{M}}$ is a topological space that is locally similar to Euclidean space~\cite{tu2008introduction}.
One can use the tangent space to model the neighbourhood structure on a differentiable manifold.
The tangent space at a point $\Mat{X}$ on the manifold, $\Mat{T_{\Mat{X}}\mathcal{M}}$, is a vector space that contains all possible directions tangentially passing through $\Mat{X}$~\cite{tu2008introduction}.

A Riemannian manifold is a differentiable manifold, endowed with a Riemannian metric.
The Riemannian metric is the family of inner products on all of the tangent spaces~\cite{jost2008riemannian}.
This metric enables us to define geometric concepts such as lengths, angles and distances.
The geodesic distance between two points $\Mat{X},\Mat{Y}$ is defined as the length of the shortest curve between
$\Mat{X}$ and $\Mat{Y}$~\cite{jost2008riemannian}.

In this section, we briefly introduce two well known Riemannian manifolds used in the computer vision community, namely Symmetric Positive Definite~(SPD) manifold and Grassmannian manifold.
\subsection{SPD Manifolds}
To compute a compact representation of an image, one method is to calculate the covariance matrix of a set of $d$-dimensional vector features extracted from the image~\cite{tuzel2006region}.
Covariance matrices naturally arise in the form of SPD matrices, which can be considered as points on SPD manifolds~\cite{tuzel2008pedestrian}. 
The geodesic distance between points on SPD manifolds then
can be calculated through an affine invariant Riemannian metric: 

\begin{equation}
\label{Eqn:dist_SPD}
\operatorname{dist}(\Mat{X},\Mat{Y})=||\operatorname{log}(\Mat{X}^{- \frac{1}{2}}\Mat{Y} \Mat{X}^{- \frac{1}{2}})||^2_F \textrm{ ,}
\end{equation}
\noindent
where $\Mat{X},\Mat{Y} \in \Mat{\mathcal{M}}$ are two points over the SPD manifold.
For further discussions on SPD manifolds, the readers are referred to~\cite{pennecetal2006}. 

To further improve clustering performance,
SPD manifolds could be projected into RKHS by Mercer kernels.
In this paper, we use one of the popular kernels for SPD manifolds, namely the Gaussian kernel, which is defined by:

\begin{equation}
\label{SPD_kernel}
\operatorname{K}(\Mat{X},\Mat{Y})=\operatorname{exp}(-\beta \cdot\operatorname{dist}(\Mat{X},\Mat{Y}))\textrm{ ,}
\end{equation}
where $\operatorname{dist}(\Mat{X},\Mat{Y})$ is the geodesic distance between point $\Mat{X}$ and $\Mat{Y}$ from Eqn.\ref{Eqn:dist_SPD}.
Since the geodesic distance is computationally demanding, several methods for computing the approximate distance have been developed~\cite{arsignyetal2006,sra2011,wang2004affine}. 
In this paper, we use two popular approximate distance functions: Log Euclidean Distance~(LED)~\cite{arsignyetal2006} and Stein Divergence~(SD)~\cite{sra2011}.
The Gaussian kernel with LED and SD then can be respectively formulated by:

\begin{equation}
\label{eqn:log_kernel}
\operatorname{K}_{LED}(\Mat{X},\Mat{Y})=\operatorname{exp}(-\beta \cdot ||\log(\Mat{X})-\log(\Mat{Y})||^2_F)
\end{equation}
and
 \begin{equation}
 \label{eqn:SD_kernel}
 \operatorname{K}_{SD}(\Mat{X},\Mat{Y})= \operatorname{exp}(-\beta \cdot \log \left( \det \left( \frac{\Mat{X}+\Mat{Y}}{2}\right) \right)
     - \frac{1}{2}  \log \left( \det \left( \Mat{X}\Mat{Y} \right) \right))\textrm{ .}
 \end{equation} 
  Note that, in order to become a Mercer kernel, the Gaussian kernel with SD requires $\beta$ to be of the form: $\beta\in\left\lbrace \frac{1}{2},\frac{2}{2},...,\frac{d-1}{2} \right\rbrace $.
\subsection{Grassmannian Manifolds}
 The Grassmannian Manifold $\Mat{\mathcal{G}_{q,d}}$, is the set of all $d$-dimensional subspaces of $\mathbb{R}^q$.
A point on the Grassmann manifold, $\Mat{X}\in \Mat{\mathcal{G}_{q,d}}$, can be denoted by an orthonormal matrix in $\mathbb{R}^{q\times d}$.
The geodesic distance between points $\Mat{X}$ and $\Mat{Y}$ on a Grassmannian manifold is defined as:

\begin{equation}
 \operatorname{dist}(\Mat{X},\Mat{Y})=\sqrt{\theta_1^2+...+\theta_d^2}\textrm{ ,}
 \end{equation}
 \noindent
 where $\theta_i$ is the principal angle between $\Mat{X}$ and $\Mat{Y}$.
 The angle $\theta_i$ can be calculated by $\theta_i=cos^{-1}(\xi_i)$ wherein $\xi_i$ are singular values of $\Mat{X}^\top\Mat{Y}$. 
 We refer readers to~\cite{absiletal2004} for further treatment on Grassmannian manifolds.
  A popular kernel used over Grassmannian manifolds is known as the Projection kernel~\cite{hammLee2008, vemulapalliPillai2013}, which can be formulated as: 
  
  \begin{equation}
  \label{eqn:G_kernel}
  \operatorname{K}(\Mat{X},\Mat{Y})=\beta \cdot||\Mat{X}^{\top}\Mat{Y}||^2_F\textrm{ .}
  \end{equation}
  

%% file: methodology.tex
\section{Proposed Framework}
\label{sec:proposed_framework}

As mentioned in Section~\ref{sec1}, the goal of our work is to significantly reduce clustering computational complexity for manifold features while maintaining reasonable clustering performance.
We address this by adopting a random projection approach to Riemannian manifolds.
In this section, we first discuss the overview of random projection in Euclidean space.
We then extend the notion into the Riemannian manifold space.

\subsection{Random Projection in Euclidean Space}
In Euclidean space, the random projection embeds original data into a much 
lower dimensional space whilst preserving the geometric structure~\cite{santosh2004}.
This can significantly reduce the computational complexity of learning algorithms, such as classification or clustering.
For instance, as a result, this is used to achieve real time performance in object tracking~\cite{salaheldin2013robust}.

A point $\Vec{x} \in \mathbb{R}^d$ in Euclidean space can be projected into a random k-dimensional subspace $(k<<d)$ via a set of randomly generated hyperplanes $\left\lbrace \Vec{r}_1\right\rbrace_{i=1}^k$ where $\Vec{r}_i \in \mathbb{R}^d$.
This can be formulated as:

\begin{equation}
\label{eqn:RP}
f(x)=\Vec{x}^\top \Mat{R}\textrm{ ,}
\end{equation}
\noindent
where $\Mat{R}$ is the random matrix that arranges the random hyperplanes as column vectors. 
Note that in order to minimise distortions produced by the projection, the matrix $\Mat{R}$ should possess a particular property.
We introduce this property in Definition~\ref{jl_projection}. 
When the random projection matrix $\Mat{R}$ possesses such a property, then the Johnson-Lindenstrauss Lemma (JL-Lemma)~\cite{johnson1984extensions} applies.
\begin{lemma}
\label{JL}[Johnson-Lindenstrauss Lemma~\cite{johnson1984extensions}]
For any $\epsilon$ such that \mbox{$ \epsilon > 0$}, and any set of points $\mathcal{X} $ with $|\mathcal{X}| = n$ upon projection to a uniform random k-dimension subspace where $k \geq \operatorname{O}(\epsilon^{-2}  \operatorname {log}\ n)$, the following property holds for every pair $\Vec{u}, \Vec{v} \in \mathcal{X}$,
	$(1-\epsilon) || \Vec{u} - \Vec{v}||^2 \leq ||\operatorname{f}(\Vec{u}) - \operatorname{f}(\Vec{v})||^2 \leq (1+\epsilon) ||\Vec{u} - \Vec{v}||^2$, where $\operatorname{f}(\Vec{u}), \operatorname{f}(\Vec{v})$ are the projections of $\Vec{u},\Vec{v}$.
\end{lemma}

\noindent
\textit{Remarks} 
The JL-Lemma principally states that a set of high dimensional points can be embedded using a set of uniform random hyperplanes into lower dimensional space wherein the pairwise distance between two points is well preserved (with high probability).
The original proof of JL-Lemma uses quite challenging geometric approximation machinery~\cite{johnson1984extensions}.
Frankl and Meahara~\cite{frankl1988johnson} simplified that proof by considering a projection into $k$
random orthonormal vectors.
Recently there have been several properties of the random matrix where JL-Lemma still applies.
We shall call the type of projection wherein the random matrix has properties that allow the JL-Lemma to be applied as a JL-Type projection.
\begin{definition}[JL-Type projection]
\label{jl_projection}
	Let $\Mat{R} = [\Vec{r}_1 \cdots \Vec{r}_k], \Vec{r}_i \in \mathbb{R}^d$ be a random matrix whose columns are the random hyperplanes. The projection $\operatorname{f}(\Vec{u}) =  \Mat{R}^{\top} \Vec{u}, \Vec{u} \in \mathbb{R}^d, \operatorname{f}(\Vec{u}) \in \mathbb{R}^k$ is called JL-Type projection when the matrix $\Mat{R}$ possesses at least one of the following properties:
	\begin{enumerate}
\item\label{orthonormal} The columns of $\Mat{R}$ are orthogonal unit-length vectors~\cite{frankl1988johnson};
\item\label{Gaussian} Each element in $\Mat{R}$ is selected independently from a standard Gaussian distribution $N(0,1)$ or uniform distribution $U(-1,1)$~\cite{arriaga1999algorithmic};
\item\label{sparse}  $\Mat{R}$ is a sparse matrix whose elements belong to $\left\lbrace -1,0,+1\right\rbrace $ with probability $\left\lbrace 1/6,2/3,1/6 \right\rbrace $~\cite{li2006very}.
\end{enumerate}
\end{definition}

\noindent
We note that Property \ref{orthonormal} in Definition~\ref{jl_projection} considers columns of the random matrix $\Mat{R}$ as the basis of a random space, thus they are required to be pairwise orthogonal~\cite{frankl1988johnson}.
To this end, one needs to apply an orthogonalisation technique such as the Gram-Schmidt method~\cite{watkins2004} on $\Mat{R}$.
Arriaga~\etal\cite{arriaga1999algorithmic} proved that it suffices to use random non-orthonormal matrices
with independent elements chosen from some distributions which are listed in Property \ref{Gaussian} of Definition~\ref{jl_projection}.
Recently, Li~\etal\cite{li2006very} proposed a
sparse random projection matrix presented in Property~\ref{sparse} of Definition~\ref{jl_projection}.
The sparse random projection achieves a further threefold speed-up as only $1/3$ of the matrix have non-zero elements.

We note that the random projection is not data driven.
It means that it does not need a set of labelled training data, making it suitable for unsupervised learning scenarios such as clustering~\cite{boutsidis2010random,sakai2009fast}. 
 
\subsection{Random Projection in Riemannian Manifolds via RKHS} 
As mentioned in Section~\ref{sec1}, applying the random projection on points residing in the Riemannian manifold space is not trivial, due to the notion of projection itself being generally not well defined.
We approach this problem by reformulating the problem in the RKHS.
Recall that, the random matrix containing column vector of hyperplanes $\Vec{r}_i$ should be generated from a particular process.
Thus, the projection of each individual dimension into the projected space is carried out as follows: 

\begin{equation}
	f_i(\Vec{x}) = \Vec{x}^\top \Vec{r}_i\textrm{ ,}
\end{equation}
\noindent
where $f_i(\cdot)$ is the $i$-th dimension of the projected vector $\Vec{x}$. 

In the RKHS, the above formulation can be rewritten as:

\begin{equation}
	f_i(\Vec{x}) = \phi(\Vec{x})^\top \Vec{r}_i\textrm{ ,}
		\label{eq:rp_in_kernel}
\end{equation}
\noindent
where $\phi(\cdot)$ is the function that embeds the input space into the RKHS. 
Note that, in this case, the hyperplane $\Vec{r}_i$ is now defined in the RKHS, $\Vec{r}_i \in \mathcal{H}$.
The projection in the RKHS can be considered as the inner product which is defined as the kernel similarity function.

\begin{figure*}
          \centerline{\includegraphics[width=\textwidth,keepaspectratio]{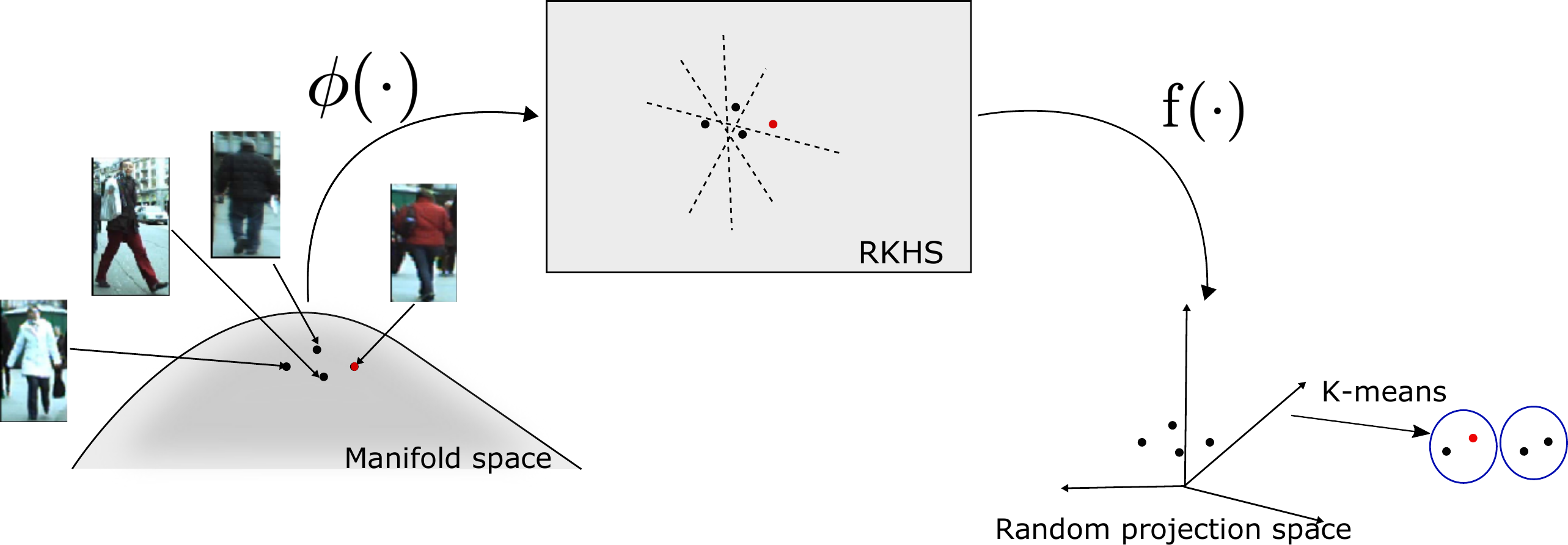}}
             \caption{
				The illustration of our proposed framework. 
				We first generate the hyperplanes in RKHS. 
				 Each point in the manifold space is then mapped into the projected space via the kernel inner product. Finally we apply K-means in the projected space. 
             }
          \label{fig:STEP1}
        \end{figure*}

Eqn.~\ref{eq:rp_in_kernel} provides insight that the JL-Type projection could be achieved
as long as one could generate the hyperplanes that follow one of the above properties in Definition~\ref{jl_projection} in the RKHS.
In similar fashion, when the data point $\Vec{x}$ is replaced by a point $\Mat{X}$ in manifold $\Mat{X} \in \Mat{\mathcal{M}}$, then one could use Eqn.~\ref{eq:rp_in_kernel} as the framework to achieve JL-Type projection in the manifold space.
As such, we propose a framework for clustering manifold points, which is briefly illustrated in Figure~\ref{fig:STEP1}.
This hyperplane generation is the central idea in our work.
First, we generate the hyperplanes over the RKHS.
The points over the manifold space are then projected into the projected space by using the specified kernel similarity function, such as the Gaussian kernel or projection kernel.
Once the manifold points have been embedded into the projected space, we apply the general K-means algorithm to perform clustering.
        
\begin{figure*}
                  \centerline{\includegraphics[width=0.90\textwidth,keepaspectratio]{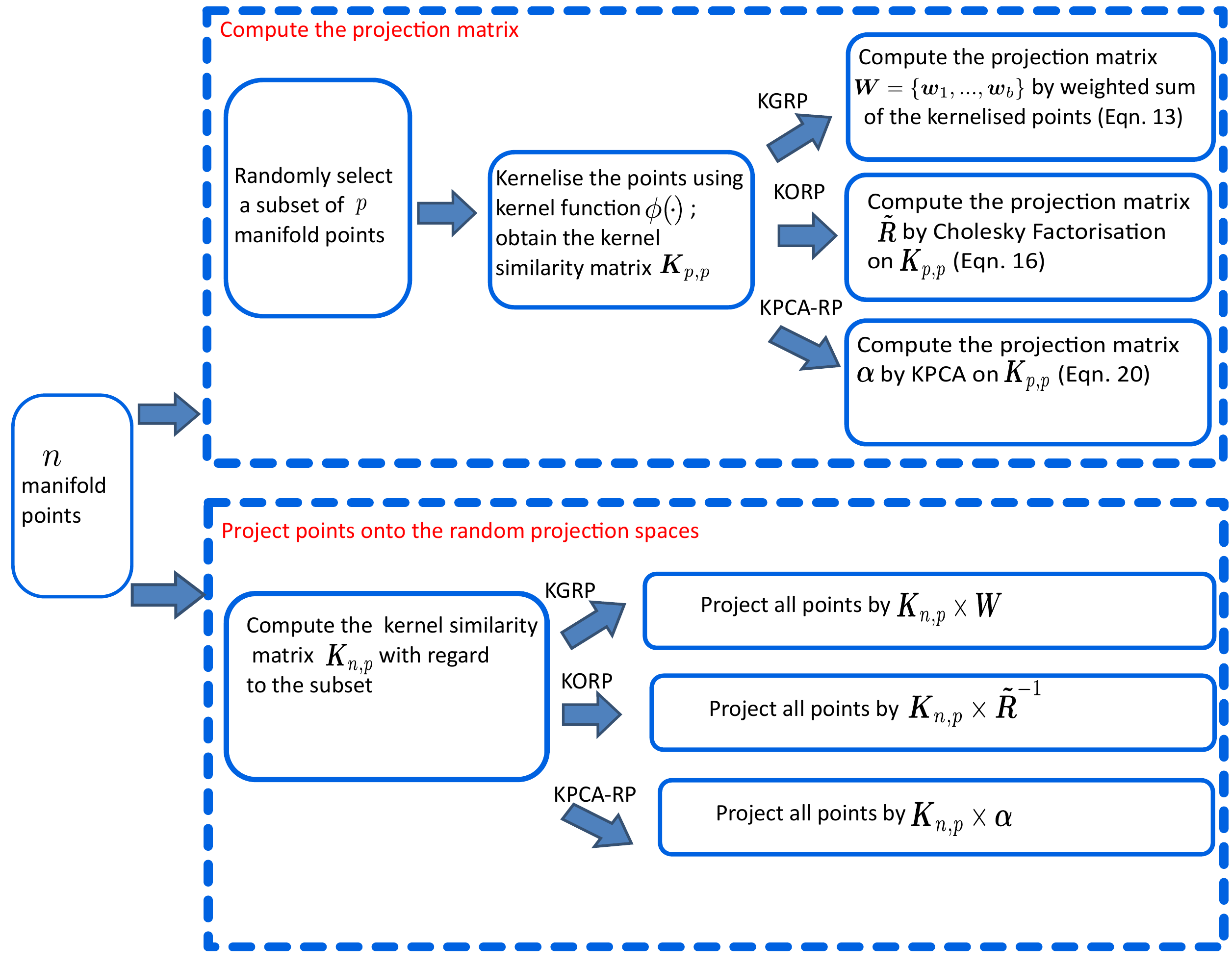}} 
                     \caption{The diagram of our proposed generation methods: KORP, KGRP and KPCA-RP.}
                  \label{fig:process}
                \end{figure*}

In this paper, we explore three hyperplane generation methods for manifold points: (1) KGRP; (2) KORP and (3) KPCA-RP.
The diagram of our proposed generation methods is illustrated in Figure~\ref{fig:process}. Briefly speaking, the hyperplanes are generated using a randomly selected subset from the entire dataset. 
The projection made by the hyperplanes will follow one of the properties in Definition~\ref{jl_projection}. We will elaborate on the generation process and theoretical analysis in the following section. 


\input{subsec_GRP}
\input{subsec_ORP}
\input{subsec_KPCARP}

%% file: subsec_GRP.tex
\subsubsection{Kernelised Gaussian Random Projection (KGRP)}
\label{sec:GRP}
In the KGRP method, the hyperplanes are generated from the standard Gaussian distribution $\mathcal{N}(\Vec{0},\Vec{I})$. 
Each hyperplane $\Vec{r}_i \in \mathcal{H}$ is assumed to be spanned by a group of data points randomly selected.
To this end, first a subset $\mathcal{S}$ 
 containing $p$ points $\{ \phi(\Mat{X}_1), \dots, \phi(\Mat{X}_p) \}$ is randomly chosen from the entire dataset, $\phi(\Mat{X}_i)$ is the representation of manifold points $\Mat{X}_i$ in the RKHS.
 Each data point $\operatorname{\phi}(\Mat{X}_i)$ from the subset is considered as a vector generated from a particular distribution $D$ with unknown mean $\Vec{\mu}$ and unknown covariance $\Mat{\Sigma}$.
 Thanks to the Central Limit
 Theorem (CLT)~\cite{rice2006mathematical}, one can still produce standard Gaussian distribution data points from these data.
 More precisely, the CLT states that when the number of data points grows larger, the difference between the population mean and the sample mean approximates the normal
  distribution $\mathcal{N}(\Vec{0}$,$\Mat{\Sigma}$).
 As such, we first randomly select t, $t < p$, data points from $\mathcal{S}$ and let these points be the set $\mathcal{S}_1 \subset \mathcal{S}$.
 Let $\Vec{z}_t = \frac{1}{t}\sum_{i\in \mathcal{S}_{1}} \operatorname{\phi}(\Mat{X}_i)$ be the sample mean over $\mathcal{S}_{1}$. 
 By applying the CLT and the Whitening transform~\cite{duda2012pattern}, the vector $\Vec{r}_i = \Mat{\Sigma} ^{-\frac{1}{2}} \sqrt t (\Vec{z}_t - \Vec{\mu})$ can be considered as the point generated from a standard Gaussian distribution; thus $\Vec{r}_i$ could be used as a random projection hyperplane. 
 Therefore, we denote our embedding function that projects data points in the RKHS to the random projection space by:
 
\begin{equation}
f({\phi}(\Vec{X}_i))=\operatorname{\phi}(\Vec{X}_i)^{T} \Mat{\Sigma} ^{-\frac{1}{2}} \sqrt t (\Vec{z}_t - \Vec{\mu})\textrm{ .}
	\label{eqn:mapping_function}
\end{equation}


\noindent
The mean is implicitly estimated as $\Vec{\mu} = \frac{1}{p}\sum_{i=1}^{p} \operatorname{\phi}(\Vec{X}_i)$, and the covariance matrix $\Mat{\Sigma}$ is also formed over the $p$ data points.
In order to compute Eqn.~\ref{eqn:mapping_function}, one could use a similar approach to that of Kernel Principal Component Analysis~(KPCA)~\cite{scholkopfetal1998}.
Specifically, let the Eigen-decomposition of the covariance matrix $\Mat{\Sigma}$ and the kernel matrix over $p$ data points $\Mat{K}_\mathcal{S}$, be $\Mat{V}\Mat{\Lambda}\Mat{V}^{\top}$ and $\Mat{U} \Mat{\Theta} \Mat{U}^{\top}$ respectively.
Based on the fact that the non-zero eigenvalues of $\Mat{V}$ are equal to the non-zero eigenvalues of $\Mat{\Theta}$,
Kulis-Grauman~\cite{kulisGrauman2012} proved that Eqn.~\ref{eqn:mapping_function} is the same as:

\noindent
\begin{equation}
	 \sum\nolimits_{i=1}^{p} \Vec{w}(i) ( {\operatorname{\phi}(\Mat{X}_i)^{T}} {\operatorname{\phi}(\Mat{X})}) 
	\label{eqn:KLSH_embedding_3}\textrm{ ,}
\end{equation}%

\noindent
where 
\begin{equation}
	\Vec{w}(i) = \frac{1}{t}\sum\limits_{j = 1}^p {\sum\limits_{l \in {\mathcal{S}_1}}^{} {{\Mat{K}_{ij}}^{ - \frac{3}{2}}} } {\Mat{K}_{jl}}\textrm{ .}
	\label{eqn:KLSH_embedding_4}
\end{equation}%
\noindent
Note that $\mathcal{S}_1$ is the set of $t$ points which are randomly selected from $\mathcal{S}$. 
Further, defining $\Vec{e}$ as a vector of all ones, and $\Vec{e}_{\mathcal{S}_{1}}$ as a zero vector with ones in the entries corresponding to the indices of $\mathcal{S}_1$,
the expression in Eqn.~\ref{eqn:KLSH_embedding_4} can be further simplified to:

\begin{equation}
	 \Vec{w} = \sqrt {\frac{{p - 1}}{t}} {\Mat{K}_{\mathcal{S}}^{ - \frac{1}{2}}}{\Vec{e}_{{\mathcal{S}_1}}}\textrm{ .}
	\label{eqn:KLSH_embedding_5}
\end{equation}%
We note that the above formulation was first described for developing the kernelise locality sensitive hashing method in Euclidean scenarios~\cite{kulisGrauman2012}.
We then adapted the method in our previous work~\cite{alavietal2014} to perform random projection on SPD manifolds for classification purposes.
Here we apply the method for clustering on Riemannian manifold problems.
The pseudo code for KGRP is summarised in Algorithm~\ref{alg:pseudocode_KGRP}. 

\begin{algorithm}[!tb]
\caption{ Kernelised Gaussian Random Projection (KGRP)}
\label{alg:pseudocode_KGRP}
\begin{algorithmic}[1]

 \REQUIRE the entire dataset: a set of manifold-valued data points $\left\lbrace \Mat{X}_{i} \right\rbrace^{n}_{i=1}$, $\Mat{X}_i \in \Mat{\mathcal{M}}$;  the size of $\mathcal{S}$~: p; the desired projected space dimensionality~: $b$
    \ENSURE $\{\Vec{x}_i\}_{i=1}^{n}$, $\Vec{x}_i \in \mathbb{R}^p$ the data points in the projected space
 \STATE Randomly select $p$ points $\left\lbrace \Mat{X}_{i} \right\rbrace^{p}_{i=1}$ from the entire dataset
     \STATE Compute the Kernel Gram matrix $\Mat{K}_{\mathcal{S}}$ over points $\left\lbrace \Mat{X}_{i} \right\rbrace^{p}_{i=1}$,
      $\Mat{K}_\mathcal{S}=\operatorname{\phi}(\Mat{X}_{{i}})^{\top}\operatorname{\phi}(\Mat{X}_{{j}})$, $\forall \Mat{X}_{{i}}, \forall \Mat{X}_{{j}}\in \left\lbrace \Mat{X}_{i} \right\rbrace^{p}_{i=1}$,
      let $\mathcal{S}=\left\lbrace \phi(\Mat{X}_{i})\right\rbrace^{p}_{i=1}$ denote the representations for these $p$ points in the RKHS 
\STATE Compute the projection matrix
 $\Mat{W}= \left\lbrace \Vec{w}_1,...,\Vec{w}_b \right\rbrace $,  $\forall \Vec{w}_i\in \mathbb{R}^p$
  \FOR{$i = 1 \to b$}
      \STATE $\mathcal{S}_1 \gets$ Randomly select $t$ data points from $\mathcal{S}$ 
	  \STATE $\Vec{e_\mathcal{S}}=\left[\Delta_1,...,\Delta_p \right]$ 
	         if $\phi(\Mat{X}_i)\in \mathcal{S}_1$, $\Delta_i=1$; otherwise $\Delta_i=0$
	
	\STATE$\Vec{w}_i = \sqrt {\frac{{p - 1}}{t}} {\Mat{K}_\mathcal{S}^{ - \frac{1}{2}}}{\Vec{e}_{{\mathcal{S}}}} $
\ENDFOR
\STATE Project each point $\Mat{X}_i$ into the random projection space: $\Vec{x}_i = \tilde{\Mat{K}} \Mat{W}$, where $\tilde{\Mat{K}}$ is the Gram matrix between $\Mat{X}_i$ and the points $\left\lbrace \Mat{X}_{i} \right\rbrace^{p}_{i=1}$

\end{algorithmic}
\end{algorithm}

We note that the total computational complexity of the KGRP algorithm is $O(np+p^3+np^2+\ell nmp)$.
Specifically, there are four factors contributing to the computational complexity:
	\begin{enumerate}
	\item
	Computing the kernel Gram matrix $\Mat{K}_{n,p}$ between $n$ points and $p$ selected points which requires $O(np)$ operations~$(p<<n)$;
	\item Generating the random hyperplanes, necessitates calculation of the kernel matrix $\Mat{K}^{-1/2}_\mathcal{S}$ for the $p$ points in $\mathcal{S}$ which requires $O(p^3)$ operations;
    \item Projecting all of the data points into the random projection space which requires $O(np^2)$ operations;
	\item Applying K-means to get the clustering results which requires $O(\ell nmp)$ operations~($\ell$ is the number of iterations of K-means, $m$ is the number of clusters and $b$ is the dimension of the projected space).
	\end{enumerate}

%% file: subsec_ORP.tex
\subsubsection{Kernelised Orthonormal Random Projection (KORP)}

In the second method, we generate orthonormal random hyperplanes (\ie the first property).
We first present the following Lemma that relates the JL-Lemma to the margin of the linear hyperplane in supervised learning settings~\cite{blum2006}.
\begin{lemma} \label{blum}
Consider any distribution over labelled examples in Euclidean space such that there exists a linear separator $\Vec{w}^{\top}\cdot \Vec{x}=0$ with margin $\lambda$. If we draw $d \geq \frac{8}{\varepsilon}\left[\frac{1}{\lambda^2}\operatorname{ln}\frac{1}{\delta} \right]$ examples $\Vec{z}_1,\cdots,\Vec{z}_d$ iid from this distribution, with probability $\geq 1-\delta$, there exists a vector $\Vec{w}'$ in span $(\Vec{z}_1,\cdots,\Vec{z}_d)$ that has error at most $\varepsilon$ at margin $\frac{\lambda}{2}$~\cite{blum2006}. 
\end{lemma}
\noindent
\textit{Proof.} 
We refer the readers to~\cite{blum2006} for the proof of this Lemma.

\noindent
\textit{Remarks.} 
Lemma~\ref{blum} essentially states that, with a high probability, the margin is still well preserved (with error at most $\varepsilon$) when the hyperplane $\Vec{w}'$ is selected from the space spanned by a subset of the data points.
Note that, as suggested in~\cite{ShiICML2012}, when the margin is well preserved, then the angle and distance between points are also well preserved.

This Lemma can also be applied for cases where the data points are in the RKHS.
This is because the RKHS is essentially an infinite-dimensional Euclidean space~\cite{blum2006}.
Given a set of points which are linearly separable with margin $\lambda$ under a particular kernel function, we draw $d$ random examples $\Vec{x}_1,\cdots,\Vec{x}_d$ from the same distribution.
 Then, according to Lemma~\ref{blum}, with probability $\geq 1-\delta$, there exists a separator in RKHS $\Vec{w}' \in \mathcal{H}$ and $\Vec{w}'=\alpha_1 \phi(\Vec{x}_1)+ \cdots +\alpha_d \phi(\Vec{x}_d)$ with error rate at most $\varepsilon$.
Note that as \mbox{$\Vec{w}'^{\top} \cdot \phi(\Vec{x})$} $=\alpha_1 \operatorname{K}(\Vec{x},\Vec{x}_1)+...+\alpha_d\operatorname{K}(\Vec{x},\Vec{x}_d)$, we then can simply consider the vector of $[\operatorname{K}(\Vec{x},\Vec{x}_1) \cdots \operatorname{K}(\Vec{x},\Vec{x}_d)]$ as the feature representation of $\Vec{x}$ in the space spanned by $\{ \phi(\Vec{x}_i) \}_{i=1}^{d}$.
In other words, the $\operatorname{K}(\Vec{x},\Vec{x}_i)$ is considered as the i-th feature of $\Vec{x}$. 
We can further formalise this observation with the following Corollary~\cite{blum2006}.
\begin{corollary}\label{cor}
If distribution $P$ has margin $\lambda$ in the RKHS, then with probability $\geq 1-\delta$, if $\Vec{x}_1,\cdots,\Vec{x}_d$ are drawn from the same distribution, for $d=\frac{8}{\varepsilon}\left[\frac{1}{\lambda^2}\operatorname{ln}\frac{1}{\delta} \right] $, the mapping $\operatorname{F_{1}}(\Vec{x})=[\operatorname{K}(\Vec{x},\Vec{x}_1) \cdots \operatorname{K}(\Vec{x},\Vec{x}_d)]$ produces a distribution $\operatorname{F_{1}}(P)$ on labelled examples in $\mathbb{R}^{d}$ that is linearly separable with error at most $\varepsilon$~\cite{blum2006}.
\end{corollary}
\noindent

\noindent
\textit{Remarks.} The above Corollary suggests the following points: (1)~one could generate random projection hyperplanes by randomly selecting a subset of data points in RKHS and then projecting a point into this space by using $\operatorname{F_{1}}(\Vec{x})$; (2)~this projection is a JL-Type projection.

In light of these facts, for our case, we randomly select $p$ points, here denote $\mathcal{S}=\{\phi(\Mat{X}_{1}),\cdots,\phi(\Mat{X}_{p})\}$ as the implicit representations of the $p$ points in RKHS.
However, as it is possible that some hyperplanes are not linearly independent, then the hyperplanes could be highly correlated.
To that end, one needs to orthogonalise the hyperplane set $\mathcal{S}$~\cite{blum2006}.
In this work, we apply QR decomposition~\cite{watkins2004} to construct a set of orthonormal basis from the original basis spanning the same subspace.
Let us arrange the original basis $\{ \operatorname{\phi}(\Mat{X}_i) \}_{i=1}^{p}$ into a matrix $\Mat{A}$.
Then the matrix $\Mat{A}$ can be decomposed into $\Mat{Q}$ and $\tilde{\Mat{R}}$ as follows:

\begin{equation}
 \Mat{A} =  [\operatorname{\phi}(\Mat{X}_{1}),\cdots,\operatorname{\phi}(\Mat{X}_{p})]=\Mat{Q}\tilde{\Mat{R}}\textrm{ ,}
\end{equation}

\noindent
where $\Mat{Q}$ is the orthonormal basis and $\tilde{\Mat{R}}$ is the upper triangular matrix.
Assuming that we have the orthonormal basis $\Mat{Q}$, then we can observe the following when a data point $\phi(\Mat{X})$ is projected into the orthonormal basis $\Mat{Q}$:

\begin{equation}
\begin{split}
\operatorname{\phi}(\Mat{X})^{\top}\Mat{Q}
&=\operatorname{\phi}(\Mat{X)}^{\top}\Mat{Q}\tilde{\Mat{R}}\tilde{\Mat{R}}^{-1} \\
&=\operatorname{\phi}(\Mat{X})^{\top} [\operatorname{\phi}(\Mat{X}_{{1}}),...,\operatorname{\phi}(\Mat{X}_{{p}})]\tilde{\Mat{R}}^{-1}\\
&=[\operatorname{\phi}(\Mat{X})^{\top}\operatorname{\phi}(\Mat{X}_{{1}}),...,\operatorname{\phi}(\Mat{X})^{\top}\operatorname{\phi}(\Mat{X}_{{p}})]\tilde{\Mat{R}}^{-1}\\
&=[\operatorname{K}(\Mat{X},\Mat{X}_{{1}}),...,\operatorname{K}(\Mat{X},\Mat{X}_{{p}})]\tilde{\Mat{R}}^{-1}\textrm{ .}
\end{split}
\label{eq:QR_proj}
\end{equation}
\noindent
In other words, one only needs to determine the upper triangular $\tilde{\Mat{R}}$ in order to do the projection.
We note that as the original basis $\{ \operatorname{\phi}(\Mat{X}_i) \}_{i=1}^{p}$ are in the RKHS then it is not trivial to apply the QR decomposition to matrix $\Mat{A}$.
To that end, we first multiply the matrix $\Mat{A}$ by its transpose.
By doing this, we will get the kernel matrix $\Mat{K}_\mathcal{S}$, where $\Mat{K}_\mathcal{S}(i,j) = \operatorname{\phi}(\Mat{X}_i)^{\top}\operatorname{\phi}(\Mat{X}_j)$, $\forall \operatorname{\phi}(\Mat{X}_i)$ and $\forall \operatorname{\phi}(\Mat{X}_j) \in \mathcal{S}$. 
Thus:

\begin{equation}
\begin{split}
	\Mat{K}_\mathcal{S} &= \Mat{A}^{\top}\Mat{A} \\
	          &= (\Mat{Q}\tilde{\Mat{R}})^{\top}\Mat{Q}\tilde{\Mat{R}} \\
	          &= \tilde{\Mat{R}}^{\top}\Mat{Q}^{\top}\Mat{Q}\tilde{\Mat{R}} \\
	          &= \tilde{\Mat{R}}^{\top}\tilde{\Mat{R}}\textrm{ .}
\end{split}	         
\end{equation}
\noindent
We can employ the Cholesky Factorisation~\cite{watkins2004} on the kernel matrix $\Mat{K}_\mathcal{S}$, in order to compute the upper triangular $\tilde{\Mat{R}}$. 
Algorithm~\ref{kop} outlines the algorithm for the proposed Kernelised Orthonormal Random Projection (KORP).

\begin{algorithm}
\caption{Kernelised Orthonormal Random Projection~(KORP)}
 \label{kop}
 \begin{algorithmic}[1] 
    \REQUIRE the entire dataset: a set of manifold-valued data points $\left\lbrace \Mat{X}_{i} \right\rbrace^{n}_{i=1}$, $\Mat{X}_i \in \Mat{\mathcal{M}}$;  the desired projected space dimensionality~: $p$
    \ENSURE $\{\Vec{x}_i\}_{i=1}^{n}$, $\Vec{x}_i \in \mathbb{R}^p$ the data points in the projected space
    \STATE Randomly select $p$ points $\left\lbrace \Mat{X}_{i} \right\rbrace^{p}_{i=1}$ from the entire dataset
          \STATE Compute the kernel Gram matrix $\Mat{K}_{\mathcal{S}}$ over points $\left\lbrace \Mat{X}_{i} \right\rbrace^{p}_{i=1}$
           $\Mat{K}_\mathcal{S}=\operatorname{\phi}(\Mat{X}_{{i}})^{\top}\operatorname{\phi}(\Mat{X}_{{j}})$, $\forall \Mat{X}_{{i}}, \forall \Mat{X}_{{j}}\in \left\lbrace \Mat{X}_{i} \right\rbrace^{p}_{i=1}$
  \STATE Apply Cholesky Factorisation to the kernel matrix $\Mat{K}_{\mathcal{S}} = \tilde{\Mat{R}}\tilde{\Mat{R}}^{\top}$
 \STATE Project each point $\Mat{X}_i$ into the random projection space: $\Vec{x}_i = \tilde{\Mat{K}} \tilde{\Mat{R}}^{-1}$, where $\tilde{\Mat{K}}$ is the Gram matrix between $\Mat{X}_i$ and the points $\left\lbrace \Mat{X}_{i} \right\rbrace^{p}_{i=1}$, 
\end{algorithmic}
 \end{algorithm}  


The computational complexity of KORP depends on the following steps:
\begin{enumerate}
\item Computing the kernel Gram matrix between the entire dataset and the subset $\mathcal{S}$ which requires $O(np)$ operations;
\item Applying Cholesky Factorisation on the kernel Gram matrix of the $p$ points in $\mathcal{S}$ which requires $O(p^3)$ operations;
\item Applying the matrix inverse of the right triangular matrix $\tilde{\Mat{R}}$ which demands $O(p^3)$ operations;
\item Projecting all of the data points into the orthonormal space with $O(np^2)$ operations; 
\item Applying K-means to get the clustering results which demands $O(\ell nmp)$ operations~($\ell$ is the number of iterations of K-means, $m$ is the number of clusters). 
\end{enumerate}
Hence, the total computational complexity is $O(np+p^3+np^2+\ell nmp)$.

%% file: subsec_KPCARP.tex
\subsection{KPCA-based Random Projection (KPCA-RP)}
\label{SEC_KPCAA-RP}
Inspired by the previous method, one can derive orthonormal projections using the Kernel PCA (KPCA).
More precisely, after generating random projection hyperplanes by randomly selecting the subset $\mathcal{S}$, one can obtain the principal components of the data points in $\mathcal{S}$ by applying the KPCA.
The principal components of $\mathcal{S}$ are then considered as the set of orthogonal random projection hyperplanes.
Finally, following Eqn.~\ref{eq:rp_in_kernel}, the entire data points can be projected into the random projection space using the hyperplanes.

Let us suppose $\Mat{C}$ is the covariance matrix of the points in $\mathcal{S}$ which have been centred:

\begin{equation}
\Mat{C}=\frac{1}{p}\sum_{i=1}^{p}{\phi}(\Mat{X}_{{i}}){\phi}(\Mat{X}_{{i}})^{\top}.
\end{equation}
To apply KPCA, one needs to solve the generalised eigen-decomposition problem:

\begin{equation}
\label{eq:PCA_V}
\tau \Vec{V}=\Mat{C}\Vec{V}\textrm{ .}
\end{equation} 
Following the same argument as KPCA~\cite{scholkopfetal1998}, the eigenvectors of the covariance matrix $C$ lie in the span of
${\phi}(\Mat{X}_{{1}}),{\phi}(\Mat{X}_{2}),..,{\phi}(\Mat{X}_{p})$:

\begin{equation}
\Vec{V}_k=\sum_{i=1}^{p}\alpha^k_{i}{\phi}(\Mat{X}_{{i}})\textrm{ ,}
\label{eq:KPCA_linearcombination}
\end{equation}
\noindent
where the set $\{ \alpha^k_i \}_{i=1}^{p}$ can be determined by solving the following equation:

\begin{equation}
p \tau \Vec{\alpha}=\Mat{K}_\mathcal{S} \Vec{\alpha}\textrm{ ,}
\label{eq:KPCA}
\end{equation}
\noindent
where $\Mat{\alpha} = [\Vec{\alpha}^1 \cdots \Vec{\alpha}^k]$ is a matrix wherein each column represents the vector $\Vec{\alpha}^k = [\alpha^k_1 \cdots \alpha^k_p]^{\top}$ whose elements are the linear combination coefficients presented in Eqn.~\ref{eq:KPCA_linearcombination} and $\Mat{K}_\mathcal{S}$ is the kernel matrix of the set $\mathcal{S}$.
Note that the above equation suggests that the vector $\Vec{\alpha}^k$ is one of the eigenvectors of $\Mat{K}_\mathcal{S}$.

Let $\{\Vec{V}_k\}_{k=1}^{p}$ be the set of principal components extracted from Eqn.~\ref{eq:PCA_V}.
To project a point into the principal component $\Vec{V}_k$, we perform:

\begin{equation}
\label{kpca_proj}
\phi(\Mat{X})^{\top} \cdot \Vec{V}_{k}=\sum_{i=1}^{p}\alpha_{i}^{k}\phi(\Mat{X})^{\top} {\phi}(\Mat{X}_{{i}})\textrm{ .}
\end{equation}
In the following, we present a theorem that guarantees that projections into the principal components of the subset $\mathcal{S}$ achieves JL-Type projection.
\begin{theorem}
If a set of points can be separated by a margin $\lambda$ in the RKHS, then with probability $\geq 1-\delta$, 
if $\mathcal{S}=\left\lbrace \phi(\Mat{X}_1),...,\phi(\Mat{X}_p)\right\rbrace$, $\Mat{X}_i \in  \Mat{\mathcal{M}}$, $\phi(\Mat{X}_i) \in \mathcal{H}$ are drawn from the same distribution for 
$p=\frac{8}{\varepsilon}\left[\frac{1}{\lambda^2}\operatorname{ln}\frac{1}{\delta} \right] $, 
the mapping $\operatorname{F_{2}}(x)=\operatorname{F_{1}}(x)[\Vec{\alpha}^1 \cdots \Vec{\alpha}^p]$, 
where $\Vec{\alpha}^k$ is the $k$-th eigenvector of $\Mat{K}_\mathcal{S}$, achieves JL-Type projection with error at most $\varepsilon$.
\label{theorem:kpcarp}
\end{theorem}
\textit{Proof.} 
As presented in Corollary~\ref{cor}, $\operatorname{F_{1}}(x)$ is the function that maps a point into a random projection space wherein the set of hyperplanes $\mathcal{S}$ is randomly selected from a set of given points. 
It is known that principal components of $\mathcal{S}$ represent the orthonormal bases spanning the subspace spanned by $\mathcal{S}$.
Henceforth, computing the principal components of $\mathcal{S}$ can be considered as orthogonalisation of the hyperplanes.

\noindent
\textit{Remarks.}
The above theorem states that applying KPCA on $\mathcal{S}$ means orthogonalising the hyperplanes in $\mathcal{S}$. 
Therefore, the difference between KPCA-RP and KORP is related to how the hyperplanes are orthogonalised.
We present the KPCA-RP pseudo code in Algorithm~\ref{kpca-rp}.




\begin{algorithm}
\caption{KPCA-based Random Projection (KPCA-RP)}
 \label{kpca-rp}
 \begin{algorithmic}[1] 
    \REQUIRE the entire dataset: a set of manifold-valued data points $\left\lbrace \Mat{X}_{i} \right\rbrace^{n}_{i=1}$, $\Mat{X}_i \in  \Mat{\mathcal{M}}$;  the desired projected space dimensionality~: $p$
    \ENSURE $\{\Vec{x}_i\}_{i=1}^{n}$ the data points in the projected space
     \STATE Randomly select $p$ points $\left\lbrace \Mat{X}_{i} \right\rbrace^{p}_{i=1}$ from the entire dataset
        \STATE Compute the kernel Gram matrix $\Mat{K}_{\mathcal{S}}$ over points $\left\lbrace \Mat{X}_{i} \right\rbrace^{p}_{i=1}$
                  $\Mat{K}_\mathcal{S}=\operatorname{\phi}(\Mat{X}_{{i}})^{\top}\operatorname{\phi}(\Mat{X}_{{j}})$, $\forall \Mat{X}_{{i}}, \forall \Mat{X}_{{j}}\in \left\lbrace \Mat{X}_{i} \right\rbrace^{p}_{i=1}$
  \STATE Apply KPCA to kernel matrix $\Mat{K}_\mathcal{S}$ to obtain the eigenvectors $\alpha$. 
 \STATE Project each point $\Mat{X}_i$ into the random projection space: $\Vec{x}_i = \tilde{\Mat{K}} \alpha$, where $\tilde{\Mat{K}}$ is the Gram matrix between $\Mat{X}_i$ and the $\left\lbrace \Mat{X}_{i} \right\rbrace^{p}_{i=1}$
\end{algorithmic}
 \end{algorithm}  

In terms of calculating the computational complexity of the KPCA-RP algorithm, one needs to consider four factors:
\begin{enumerate}
\item Computing the kernel Gram matrix between the entire dataset and the subset $\mathcal{S}$, which requires $O(np)$ operations;
\item Applying KPCA on the kernel Gram matrix of subset $\mathcal{S}$, which requires $O(p^3)$ operations;
\item Projecting all of the data points into the orthonormal space, which requires $O(np^2)$ operations;
\item Applying K-means to get the clustering results, which requires $O(\ell nmp)$ operations~($\ell$ is the number of iterations of K-means, $m$ is the number of clusters). 
\end{enumerate}

Hence, the total computational complexity is $O(np+p^3+np^2+\ell nmp)$.

%% file: results.tex
\section{Experimental Results}
\label{sec:experimental_results}

We evaluate our proposal using \rev{six} benchmark datasets: (1)~Ballet dataset~\cite{wang2009human};
(2)~UCSD traffic dataset~\cite{chan2005probabilistic};
(3)~UCF101 Human actions dataset~\cite{soomro2012ucf101};
\rev{(4)}~Brodatz texture dataset~\cite{randenHusoy1999};~\rev{(5)}~KTH-TIPS2b material dataset~\cite{caputoetal2005}
and \rev{(6)}~HEp-2 Cell ICIP2013 dataset~\cite{foggia2013benchmarking}.

In our evaluation, we consider each video of the first three datasets (\ie Ballet, UCSD and UCF101) as an image set which can be effectively modelled as a point on Grassmannian manifolds. 
In addition, we use SPD manifold to model images of the latter three datasets (\ie Brodatz, KTH-TIPS2b and HEp-2 Cell ICIP2013).
To demonstrate the efficacy of our framework, we report the clustering performance and the run time.
\subsection{Datasets and Feature Extraction}
\begin{figure}
 \centering
 \begin{subfigure}[b]{0.3\textwidth}
   \includegraphics[width=3.5cm,height=2cm]{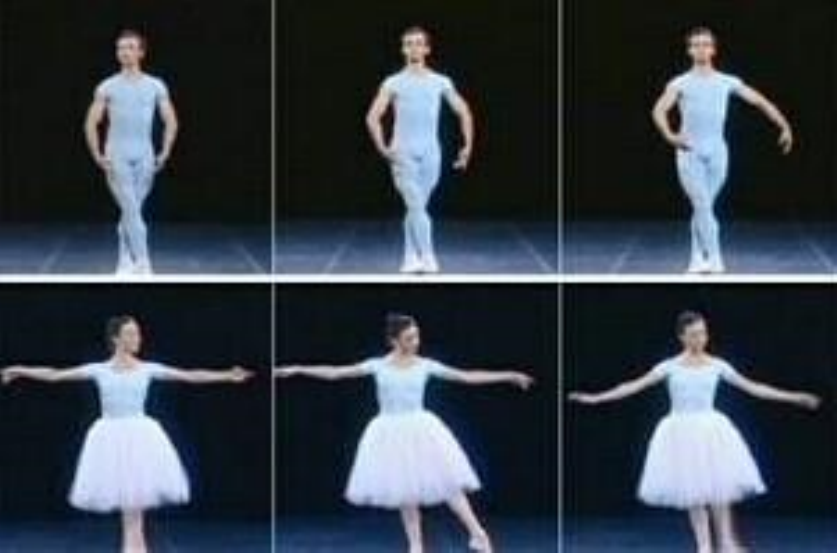}
  \caption{ 
   }
   \label{fig:Ballet}
   \end{subfigure}
   ~~
 \begin{subfigure}[b]{0.3\textwidth}
                \includegraphics[width=3.5cm,height=2cm]{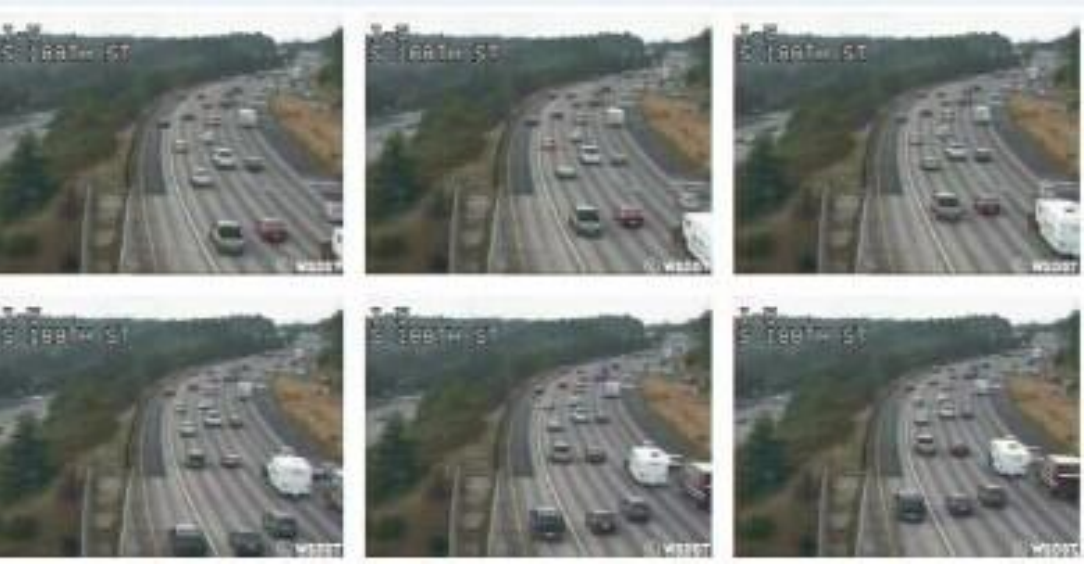}
                \caption{}
                \label{fig:UCSD}
        \end{subfigure}
        \begin{subfigure}[b]{0.3\textwidth}
           \includegraphics[width=3.5cm,height=2cm]{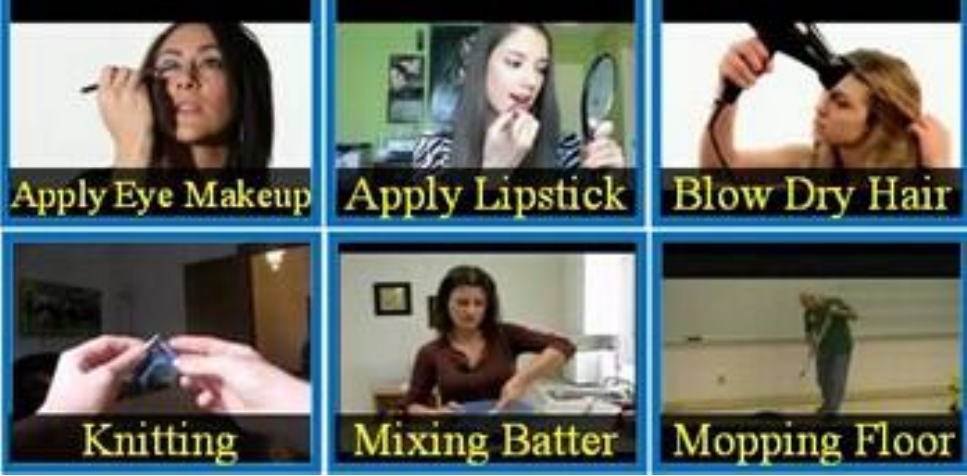}
          \caption{ 
           }
           \label{fig:ucf}
           \end{subfigure}
         \caption{Examples from~(a)~Ballet action dataset~\cite{wang2009human}~(b)~UCSD traffic dataset~\cite{chan2005probabilistic} and (c)~UCF101 dataset~\cite{soomro2012ucf101}}
         \label{fig:dataset1} 
        \end{figure}

\noindent
\textbf{Ballet action dataset (Ballet)~\cite{wang2009human} -}
The Ballet dataset presents sequences of videos of ballet actions.
More precisely, it comprises 44 sequences with 8 different actions: R-L presenting, L-R presenting,
Presenting, Jump \& swing, Jump, Turn, Step, and Stand still~(see Figure~\ref{fig:Ballet} for examples).
These ballet actions were performed by two men and one woman, resulting in significant intra-class variations such as speed, clothing and movements.
In this evaluation, each video is considered as an image set.
We then represent each image set as a point in the Grassmannian manifold.
To that end, all the videos are down sampled to $16 \times 16$ pixels.
A Grassmannian point is extracted for every $6$ consecutive frames.
Technically, we first vectorise each frame into a column vector and arrange them into a $256 \times 6$ tall matrix (\ie $256 = 16 \times 16$).
The matrix can be considered as a subspace and the orthonormal bases spanning the subspace can be determined by applying the Singular Value Decomposition (SVD).
The set of orthonormal bases is considered as a Grassmannian point~\cite{shirazietal2012}.
We use the projection kernel~(see Eqn.~\ref{eqn:G_kernel}) in this evaluation.

\noindent

\noindent
\textbf{UCSD traffic dataset (UCSD)~\cite{chan2005probabilistic} - }
The UCSD traffic dataset consists of 254 video sequences
collected from the highway traffic over two days in Seattle~(see Figure~\ref{fig:UCSD} for examples).
It contains a variety of traffic patterns and weather conditions
~(\ie overcast, raining, sunny).
In total, there
are 44 sequences of heavy traffic (slow, stop and go
speeds), 45 sequences of medium traffic (reduced speed), and 165 sequences of
light traffic (normal speed).
To extract a Grassmannian point, we first randomly select half the number of frames from each video.
Each frame in each sequence is downsized to $140\times161$ pixels and further normalised by subtracting the mean frame and dividing the variance.
Then, we apply the two dimensional Discrete Cosine Transform (DCT) on the frame and use the DCT coefficients as the feature vector for each frame.
SVD is applied on the feature vectors of the frames to obtain the set of orthonormal bases.
We also choose the projection kernel~(see Eqn.~\ref{eqn:G_kernel}) for this dataset.

\noindent
\textbf{UCF101 Human Actions dataset (UCF101)~\cite{soomro2012ucf101} - }
This dataset consists of $13,320$ videos that belong to 101 categories.
For example, Applying \rev{E}ye \rev{M}akeup, Blow Dry Hair and Mixing Batter (refer to Figure~\ref{fig:ucf}).
For each video, we first extract the normalised pixel intensities as features for all the frames.
Then the SVD is applied on these features of each video to obtain the Grassmannian manifold point.
Thus, in this dataset, there are $13,320$ manifold points in total. Projection kernel (see Eqn.~\ref{eqn:G_kernel}) is used.

\begin{figure}[!b]
                \centering
                \begin{subfigure}[b]{0.3\textwidth}
                        \includegraphics[width=3.5cm,height=1.5cm]{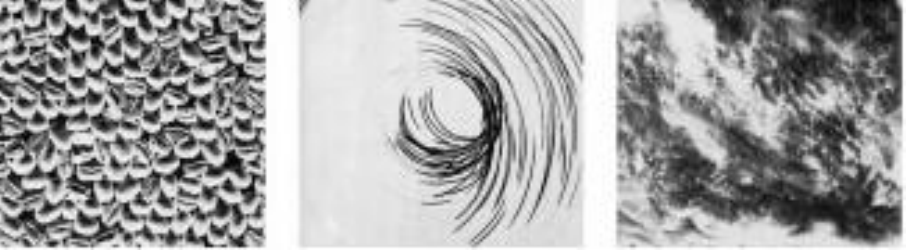}
                        \caption{}
                        \label{fig:BRODATZ_examples}
                \end{subfigure}
                ~~~~
                \begin{subfigure}[b]{0.3\textwidth}
                        \includegraphics[width=3.5cm,height=1.5cm]{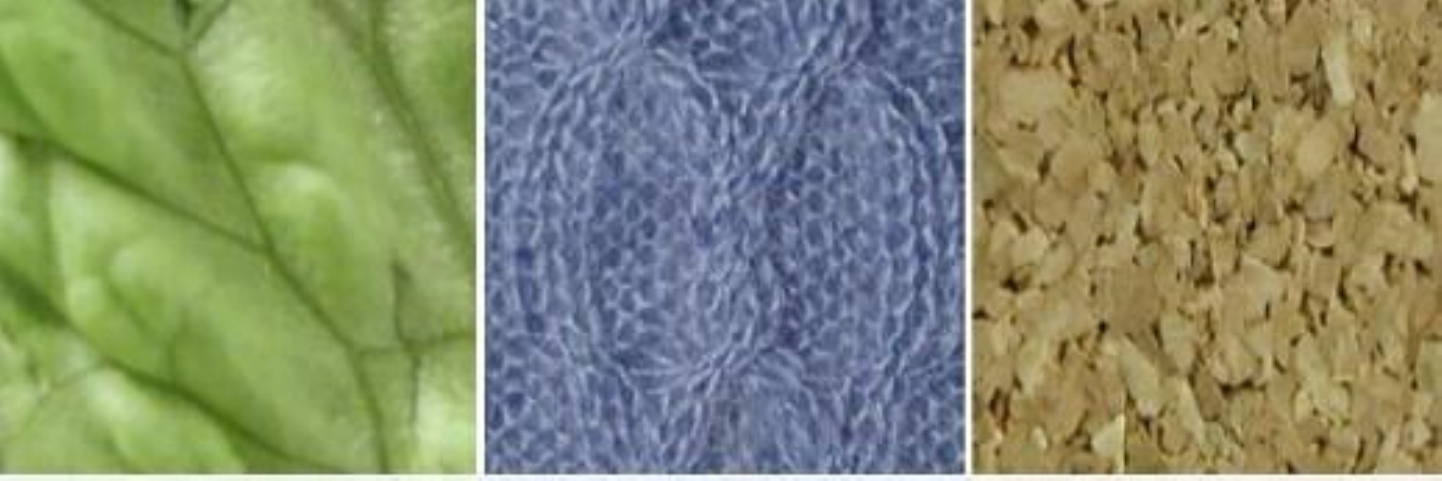}
                        \caption{}
                        \label{fig:KTH-TIPS2b}
                \end{subfigure}
                ~~~~\begin{subfigure}[b]{0.3\textwidth}
                                       \includegraphics[width=3.5cm,height=1.5cm]{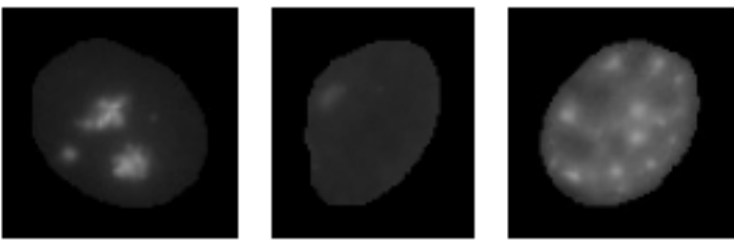}
                                       \caption{}
                                       \label{fig:cell_examples}
                               \end{subfigure}
                \caption{Examples from~(a)~BRODATZ texture dataset~\cite{randenHusoy1999},~(b)~KTH-TIPS2b material dataset~\cite{caputoetal2005} and (c)~HEp-2 Cell ICIP2013 dataset~\cite{foggia2013benchmarking}.}
                \label{fig:dataset}
        \end{figure}

\noindent
\textbf{Brodatz texture dataset (Brodatz)~\cite{randenHusoy1999} - } 
For the Brodatz dataset~(refer to Figure~\ref{fig:BRODATZ_examples} for examples)   
we follow the protocol presented in~\cite{sivalingametal2010}.
The protocol
includes 3 subsets with different numbers of classes: 5-class-texture (‘5c’, ‘5m’, ‘5v’, ‘5v2’, ‘5v3’), 10-class-texture
(‘10’, ‘10v’) and 16-class-texture (‘16c’, ‘16v’).
Each image is down-sampled to $256\times 256$ pixels and divided into 64 $32\times 32$ pixel size regions.
A feature vector $F(x, y)$ for each pixel is calculated using 
the grayscale intensities and absolute values of the first- and second-order 
derivatives of spatial feature vectors.
It can be illustrated as:
$
F(x, y)
  \mbox{~=~}
  \left[
    I\left(x,y\right),
    \left| \frac{\partial I}  {\partial x}  \right|,  \left|\frac{\partial I}  {\partial y}  \right|,
    \left| \frac{\partial^2 I}{\partial x^2}\right|,  \left|\frac{\partial^2 I}{\partial y^2}\right|
  \right] \nonumber
$.
Each region is represented by a covariance matrix~(SPD matrix)~formed 
from these feature vectors.
The Gaussian Kernel with Log-Euclidean distance~(see Eqn.~\ref{eqn:log_kernel}) is used for this dataset.

\noindent
\textbf{KTH-TIPS2b material dataset (KTH-TIPS2b)~\cite{caputoetal2005} - } This dataset contains 11 material categories captured under 4 different illuminations, in 3 poses and at 9 scales~(refer to Figure~\ref{fig:KTH-TIPS2b}).
Thus, there are $3\times4\times9=108$ images for each sample in one category, with $4$ samples per material.
We extract a 20-dimensional feature vector for each pixel in the image:
\begin{equation}
\label{eq:cov_features}
	[ \Mat{I}(x,y), \Mat{Y}(x,y), \Mat{C}_b(x,y), \Mat{C}_r{(x,y)}, F^1_{(x,y)}(\Mat{Y}) \cdots F^{16}_{(x,y)}(\Mat{Y})],	
\end{equation}
\noindent
where $\Mat{I}(x,y)$ is the image grey level value at location $(x,y)$; $\Mat{Y}$, $\Mat{C}_b$ and $\Mat{C}_r$ are the perceptually uniform CIELab colour space; The filter banks $F^i$ consist of different of offset Gaussians applied on the luminance channel $\Mat{Y}$~\cite{tosatoetal2013}.
The covariance matrix is computed once the feature vectors are extracted from every pixel location.
This becomes the image representation over a SPD manifold.
For the manifold kernel in this dataset, we use Gaussian kernel with the Stein Divergence~(see Eqn.~\ref{eqn:SD_kernel}) as this has been shown to be effective in various classification problem domains~\cite{alavietal2014, alavi2014multi}. 

\noindent
\textbf{HEp-2 Cell ICIP2013 dataset~\cite{foggia2013benchmarking} - } 
This dataset contains $13,596$ cell images that include six cell patterns namely Centromere, Golgi, Homogeneous, Nucleolar, Nuclear Membrane, and Speckled (refer to Figure~\ref{fig:cell_examples}).
The cell boundary of every cell image is described by a mask image of the same size.
For each cell image, we first extract the following feature vector of each pixel that belongs to the cell content: 
$
F(x, y)
  \mbox{~=~}
  \left[
    \left| \frac{\partial I}  {\partial x}  \right|,  \left|\frac{\partial I}  {\partial y}  \right|,
    I\left(x,y\right), 
    \left| \frac{\partial^2 I}{\partial x^2}\right|,  
    \left|\frac{\partial^2 I}{\partial y^2}\right|
  ,
  \operatorname{arctan}( \left| \frac{\partial I}  {\partial x}  \right| /\left|\frac{\partial I} {\partial y}  \right|)\right]
   \nonumber
$. 
Then, the covariance matrix~(SPD matrix) is formed 
from these feature vectors extracted from each image.
We also use Gaussian kernel with the Stein Divergence~(see Eqn.~\ref{eqn:SD_kernel}) for the evaluation on this dataset.

  
  

\subsection{Experimental Settings}
\label{sec:ex_setting}
As illustrated in Figure~\ref{fig:STEP1}, we first randomly project the points and then apply K-means.
As such, for each dataset, we first run each proposed projection method 10 times to generate 10 different random projection representations.
Then, for each representation, we run the K-means algorithm 10 times, resulting in each method being repeated 100 times for each evaluation.
The average of clustering performance and run time were reported.
As the source of variation for the other approaches is predominantly on the initial cluster seeds of K-means, we only repeat the experiment 10 times to obtain the average clustering performance and run time.

All of the approaches are tuned to give the best performance. 
We find the optimum size of set $\mathcal{S}$ as follows: (1)~Ballet: 100; (2)~UCSD: 90; (3)~UCF101: 101;~(4)~Brodatz: 100; (5)~KTH-TIPS2b: 48 and~(6)~HEp-2 Cell ICIP2013: 60.
In addition, for KGRP, we set the number of dimensionality, $b$, to 300. 
  

To measure the clustering quality, there are two main types of metrics: internal metrics based on the \rev{distances} between data points in the space, and external metrics based on the labels of the data~\cite{manningetal2008}.
The clustering task in our proposed framework is performed in a transformed space which may have different scale to other spaces used by \rev{comparable} methods such as LogE \rev{(see below for further discussion on LogE)}. This may make the internal metrics such as Dunn Index unsuitable in our case. Thus, we choose four external metrics to measure the clustering quality: Rand Index (RI), Cluster Purity (CP), F-Measure and Normalized Mutual Information~(NMI). Interested readers are referred to~\cite{manningetal2008} for further explanation of each metric.
In addition, 
we also measure the run time (in seconds) of each approach on every evaluation.
The run time is measured from the kernel matrix 
computation until the completion of the clustering process.
Finally, we report the average run time of the approaches.

Our proposal is contrasted to six approaches: 
(1)~Intrinsic K-means~\cite{Turagaetal2011}; (2)~Log-Euclidean K-means~\cite{faraki2014fisher}; (3)~Kernel K-means~\cite{dhillon2004kernel,jayasumanaetal2013}; 
(4)~KPCA K-means~\cite{scholkopfetal1998,jayasumanaetal2013}; (5)~Sigma set K-means~\cite{hongetal2009} \rev{and}
(6)~Grassmanian clustering~\cite{shirazietal2012}. 
The following is the brief description of each approach.

\noindent
{\bf Intrinsic K-means (Intrinsic):} To cluster a set of manifold points, Intrinsic K-means works directly on the manifold space using the appropriate geodesic distance~\cite{Turagaetal2011}. We note that as the intrinsic approach is generally very slow, we stop the Intrinsic K-means after 100 iterations.

\noindent
{\bf Log-Euclidean K-means~(LogE):} We first project all of the manifold points into the tangent space at the identity~\cite{arsignyetal2006}.
Once projected, each point will be vectorised into a column vector.
As for SPD manifolds, we follow the work in~\cite{pennecetal2006} that uses only the upper triangular elements.
This trick will reduce the final representation dimensionality, markedly reducing the run time on the subsequent process.
Unfortunately the trick cannot be used on Grassmannian manifolds since the representation for a point on the Grassmannian manifold is not a symmetric matrix.
In this case, all the elements are used in the final representation.
This could adversely affect the overall run time when the manifold dimensionality is high.
In the final step, K-means algorithm is applied.
Log-Euclidean k-means has been used for clustering large amount of manifold data~\cite{faraki2014fisher}.

\noindent
{\bf Kernel K-means:} This approach embeds manifold points into RKHS. 
Then Kernel K-means is applied to perform clustering~\cite{dhillon2004kernel,jayasumanaetal2013}.

\noindent
{\bf KPCA K-means (KPCA)}: 
All manifold points are first embedded into RKHS. 
Then, KPCA 
is used for projecting the points in RKHS into the space spanned by the principal components~\cite{scholkopfetal1998,jayasumanaetal2013}.
Finally, the K-means is applied.

\noindent
{\bf Sigma set K-means~(SIS):} Hong~\etal\cite{hongetal2009} proposed a novel descriptor for SPD manifolds which simplifies the computations of distance and mean.
Using their proposed descriptors, we apply K-means with novel efficient computations of mean and distance.

\noindent
{\bf Grassmanian clustering~(G-clustering)}
Shirazi~\etal\cite{shirazietal2012} proposed a clustering method for Grassmanian manifolds which use the eigenvectors of the normalised projection kernel matrix as the new features of Grassmanian points.

\subsection{Comparative Analysis on Clustering Quality}

\begin{table}
  \centering
  \caption
    { The clustering quality with variance (in \%) measured by Rand Index (RI), Cluster Purity (CP), F-Measure and Normalized Mutual Information~(NMI) on Ballet dataset. The best performance is in bold.
    We refer to Section~\ref{sec:ex_setting} for further explanation of each approach.}

    \label{tab:ballet}
    \vspace{0.5ex}
    \scalebox{1.0}{
    \begin{tabular}{c|cccc}
     \toprule
        {\bf Methods/Measurements}
     &{\bf RI}
     &{\bf CP}
     &{\bf F-Measure}
     &{\bf NMI}\\
        \midrule
      {\bf Intrinsic~\cite{Turagaetal2011}}~&$73.68{\pm0.00} $&$34.92{\pm0.00}$&${33.81\pm0.00}$&${21.73\pm0.00}$\\
            
        
          {\bf LogE~\cite{faraki2014fisher}}
          &$78.23{\pm 0.15}$&$20.85 {\pm2.66}$&${14.81\pm0.37 }$&${3.91\pm0.81}$\\ 
         	 {\bf G-clustering~\cite{shirazietal2012}}&$76.41{\pm0.07}$&$18.63{\pm0.58}$&${16.39\pm0.25}$&${3.51\pm0.47}$\\ 
             {\bf Kernel K-means~\cite{dhillon2004kernel,jayasumanaetal2013} }~&$\bf{79.89{\pm0.80}}$&$ 40.86{\pm3.06}$&${32.92\pm3.21}$&${32.00\pm2.73}$\\ 
             {\bf KPCA~\cite{scholkopfetal1998,jayasumanaetal2013} }~&$78.62{\pm 2.14}$&$42.30{\pm 3.33}$&${36.27\pm2.68}$&${{34.80\pm3.48}}$\\ 
            \bottomrule
             {\bf KGRP}&$78.02{\pm 1.79}$&$41.89{\pm2.43}$&${37.98\pm2.79}$&${34.05\pm2.41}$\\ 
             {\bf KORP}&$78.28 {\pm1.68}$&$\bf{42.54{\pm2.37}}$&$\bf{{38.68\pm2.81}}$&$\bf{{35.30\pm2.80}}$\\ 
             {\bf KPCA-RP}&$77.81{\pm1.94}$&$41.90{\pm2.31}$&${{38.23\pm3.11}}$&${34.64\pm2.75}$\\ 
\bottomrule
\end{tabular}
}
\end{table}

\begin{table}
  \centering
  \caption
    {
   The clustering quality with variance (in \%) measured by Rand Index (RI), Cluster Purity (CP), F-Measure and Normalized Mutual Information~(NMI) on UCSD dataset. The best performance is in bold.
      We refer to Section~\ref{sec:ex_setting} for further explanation of each approach.
    } 
    \label{tab:UCSD}
    \vspace{0.5ex}
    \scalebox{1.0}{
    \begin{tabular}{c|cccc}
     \toprule
        {\bf Methods/Measurements}
     &{\bf RI}
     &{\bf CP}
     &{\bf F-Measure}
     &{\bf NMI}\\
        \midrule
      {\bf Intrinsic~\cite{Turagaetal2011}}~&$73.26{\pm0.00} $&$74.70{\pm0.00}$&$\bf{{75.15\pm0.00}}$&${36.18\pm0.00}$\\
            
        
          {\bf LogE~\cite{faraki2014fisher}}
          &$55.24{\pm 3.25}$&$67.23 {\pm2.66}$&${40.39\pm2.81 }$&${19.11\pm3.59}$\\ 
         	 {\bf G-clustering~\cite{shirazietal2012}}&$50.68{\pm0.11}$&$64.82{\pm0.00}$&${34.31\pm0.12}$&${0.92\pm0.29}$\\ 
             {\bf Kernel K-means~\cite{dhillon2004kernel,jayasumanaetal2013} }~&$69.98{\pm7.06}$&$ 77.96{\pm4.77}$&${57.34\pm10.22}$&${45.50\pm9.71}$\\ 
             {\bf KPCA~\cite{scholkopfetal1998,jayasumanaetal2013} }~&$\bf{77.90{\pm 5.97}}$&$80.08{\pm 2.96}$&$\bf{{69.29\pm7.56}}$&$\bf{{51.31\pm6.09}}$\\ 
            \bottomrule
             {\bf KGRP}&$75.61{\pm 3.48}$&$79.64{\pm2.07}$&${66.97\pm5.17}$&${48.29\pm3.80}$\\ 
             {\bf KORP}&$77.25 {\pm1.25}$&$\bf{80.18{\pm0.74}}$&${{68.99\pm1.62}}$&${{50.58\pm1.67}}$\\ 
             {\bf KPCA-RP}&$76.46{\pm2.79}$&$79.64{\pm1.68}$&${68.60\pm3.50}$&${49.74\pm3.02}$\\ 
\bottomrule
\end{tabular}
}
\end{table}

\begin{table}
  \centering
  \caption
    {
    The clustering quality with variance (in \%) measured by Rand Index (RI), Cluster Purity (CP), F-Measure and Normalized Mutual Information~(NMI) on UCF101 dataset. The best performance is in bold.
       We refer to Section~\ref{sec:ex_setting} for further explanation of each approach.
    } 
    \label{tab:UCF101}
    \vspace{0.5ex}
    \scalebox{1.0}{
    \begin{tabular}{c|ccc|ccc}
     \toprule
        {\bf Methods/Measurements}
     &{\bf RI}
     &{\bf CP}
     &{\bf F-Measure}
     &{\bf NMI}\\
        \midrule
      {\bf Intrinsic~\cite{Turagaetal2011}}~&${97.53\pm0.00} $&${12.94\pm0.00}$&${7.43\pm0.00}$&${27.65\pm0.00}$\\

          {\bf LogE~\cite{faraki2014fisher}}
          &${97.89\pm 0.02}$&$ {8.21\pm0.15}$&${3.62\pm0.06 }$&${18.68\pm0.07}$\\ 
             {\bf Kernel K-means~\cite{dhillon2004kernel,jayasumanaetal2013} }~&${97.71\pm0.06}$&$ {15.97\pm0.48}$&${8.80\pm0.35}$&${32.35\pm0.31}$\\ 
             {\bf KPCA~\cite{scholkopfetal1998,jayasumanaetal2013} }~&${97.69\pm0.02 }$&$\bf{{17.66\pm0.33 }}$&$\bf{{9.47\pm0.19}}$&$\bf{{33.66\pm0.18}}$\\ 
            \bottomrule
             {\bf KGRP}&$\bf{{97.90\pm0.03 }}$&${15.38\pm0.28}$&${7.40\pm0.15}$&${30.96\pm0.21}$\\ 
             {\bf KORP}&$\bf{97.90\pm0.02}$&${15.69\pm0.28}$&${7.62\pm0.17}$&${31.47\pm0.17}$\\ 
             {\bf KPCA-RP}&${97.89\pm0.03}$&${15.66\pm0.33}$&${7.59\pm0.17}$&${31.38\pm0.23}$\\ 
\bottomrule
\end{tabular}
}
\end{table}

\begin{table}
  \centering
  \caption
    {
    The clustering quality with variance (in \%) measured by Rand Index (RI), Cluster Purity (CP), F-Measure and Normalized Mutual Information~(NMI) on BRODATZ dataset. The best performance is in bold.
       We refer to Section~\ref{sec:ex_setting} for further explanation of each approach.
    } 
    \label{tab:BRODATZ}
    \vspace{0.5ex}
    \scalebox{1.0}{
    \begin{tabular}{c|cccc}
     \toprule
        {\bf Methods/Measurements}
     &{\bf RI}
     &{\bf CP}
     &{\bf F-Measure}
     &{\bf NMI}\\
        \midrule
      {\bf Intrinsic~\cite{Turagaetal2011}}~&$92.29{\pm0.00} $&$79.05{\pm0.00}$&${74.20\pm0.00}$&${75.94\pm0.00}$\\
            
        {\bf SIS~\cite{hongetal2009} }&$91.42{\pm0.00}$&$76.99{\pm0.00}$&${69.68\pm0.00}$&${72.84\pm0.00}$\\ 
        
          {\bf LogE~\cite{faraki2014fisher}}
          &$92.04{\pm0.78}$& $78.34{\pm2.34}$&${74.10 \pm2.10}$&${76.13\pm1.45}$\\ 
             {\bf Kernel K-means~\cite{dhillon2004kernel,jayasumanaetal2013} }~&$93.15{\pm0.95}$&$81.40{\pm2.75}$&${75.62\pm2.13}$&${78.19\pm1.83}$\\ 
             {\bf KPCA~\cite{scholkopfetal1998,jayasumanaetal2013} }~&$\bf{93.89{\pm0.22}}$&${82.60{\pagebreak\pm1.14}}$&$\bf{{76.64\pm0.66}}$&$\bf{{79.44\pm0.57}}$\\ 
            \bottomrule
             {\bf KGRP}&$93.47{\pm0.78}$&$82.22{\pm2.34}$&${75.84\pm1.82}$&${78.49\pm1.49}$\\ 
             {\bf KORP}&$93.66 {\pm0.77}$&$82.58{\pm2.32}$&${76.30\pm1.81}$&${79.11\pm1.50}$\\
             {\bf KPCA-RP}&$93.77{\pm0.84}$&$\bf{82.81{\pm2.49}}$&${76.39\pm1.93}$&${79.16\pm1.56}$\\ 
\bottomrule
\end{tabular}
}
\end{table}

\begin{table}
  \centering
  \caption
    {
    The clustering quality with variance (in \%) measured by Rand Index (RI), Cluster Purity (CP), F-Measure and Normalized Mutual Information~(NMI) on KTH-TIPS2b dataset. The best performance is in bold.
          We refer to Section~\ref{sec:ex_setting} for further explanation of each approach.
    } 
    \label{tab:KTH-TIPS2b}
    \vspace{0.5ex}
    \scalebox{1.0}{
    \begin{tabular}{c|cccc}
     \toprule
        {\bf Methods/Measurements}
     &{\bf RI}
     &{\bf CP}
     &{\bf F-Measure}
     &{\bf NMI}\\
        \midrule
      {\bf Intrinsic~\cite{Turagaetal2011}}~&$86.99{\pm0.00} $&$49.45{\pm0.00}$&${36.19\pm0.00}$&${44.20\pm0.00}$\\
            
        {\bf SIS~\cite{hongetal2009} }&$80.81{\pm0.00}$&$41.62{\pm0.00}$&$\bf{{44.45\pm0.00}}$&${40.47\pm0.00}$\\ 
        
          {\bf LogE~\cite{faraki2014fisher}}
          &$85.94{\pm0.60 }$&$45.19{\pm1.32}$&${33.48\pm1.01 }$&${40.69\pm0.82}$\\ 
         	
             {\bf Kernel K-means~\cite{dhillon2004kernel,jayasumanaetal2013} }~&$88.35{\pm0.35}$&$ 52.59{\pm1.37}$&${41.11\pm1.01}$&$\bf{{51.08\pm0.82}}$\\ 
             {\bf KPCA~\cite{scholkopfetal1998,jayasumanaetal2013} }~&$\bf{88.48}{\pm 0.40}$&$\bf{53.38{\pm 1.53}}$&${{41.22\pm1.35}}$&${50.97\pm0.90}$\\ 
            \bottomrule
             {\bf KGRP}&$88.41{\pm 0.42}$&$53.15{\pm1.34}$&${40.48\pm0.99}$&${49.87\pm1.01}$\\ 
             {\bf KORP}&${88.36}{\pm0.39}$&$53.04{\pm1.10}$&${40.61\pm0.93}$&${50.06 \pm0.92}$\\ 
             {\bf KPCA-RP}&$88.35{\pm0.44}$&$53.45{\pm1.35}$&${40.21\pm0.91}$&${49.97\pm1.09}$\\ 
\bottomrule
\end{tabular}
}
\end{table}
\begin{table}[!t]
  \centering
  \caption
    {
   The clustering quality with variance (in \%) measured by Rand Index (RI), Cluster Purity (CP), F-Measure and Normalized Mutual Information~(NMI) on HEp-2 Cell ICIP2013 dataset. The best performance is in bold.
             We refer to Section~\ref{sec:ex_setting} for further explanation of each approach.
    } 
    \label{tab:cell}
    \vspace{0.5ex}
    \scalebox{1.0}{
    \begin{tabular}{c|cccc}
     \toprule
        {\bf Methods/Measurements}
     &{\bf RI}
     &{\bf CP}
     &{\bf F-Measure}
     &{\bf NMI}\\
        \midrule
      {\bf Intrinsic~\cite{Turagaetal2011}}~&${73.96\pm0.00} $&${44.02\pm0.00}$&${35.69\pm0.00}$&${22.65\pm0.00}$\\
            
        {\bf SIS~\cite{hongetal2009} }&${74.50\pm0.00}$&${39.50\pm0.00}$&${27.32\pm0.00}$&${18.01\pm0.00}$\\ 
       
          {\bf LogE~\cite{faraki2014fisher}}
          &${74.80\pm0.95}$&${46.00 \pm2.37}$&${34.75\pm0.86 }$&${23.64\pm1.29}$\\ 
              {\bf Kernel K-means~\cite{dhillon2004kernel,jayasumanaetal2013}}&${73.96\pm2.14}$&${46.45\pm3.30}$&$\bf{{37.29\pm3.29}}$&${24.29\pm1.95}$\\ 
             {\bf KPCA~\cite{scholkopfetal1998,jayasumanaetal2013} }~&$\bf{{75.74\pm 2.87}}$&${48.48\pm1.94 }$&${34.23\pm2.20}$&${25.29\pm0.00}$\\ 
            \bottomrule
             {\bf KGRP}&${75.72\pm0.31 }$&$\bf{{49.05\pm1.10}}$&${34.83\pm0.84}$&$\bf{{25.74\pm0.82}}$\\
             {\bf KORP}&${75.63 \pm0.62}$&${48.70\pm2.34}$&${34.73\pm1.67}$&${25.49\pm1.72}$\\ 
             {\bf KPCA-RP}&${75.72\pm0.41}$&${48.70\pm2.56}$&${34.48\pm1.74}$&${25.46\pm1.93}$\\ 
\bottomrule
\end{tabular}
}
\end{table}

Tables~\ref{tab:ballet}, \ref{tab:UCSD},~\ref{tab:UCF101},~\ref{tab:BRODATZ},~\ref{tab:KTH-TIPS2b} and~\ref{tab:cell} report the average clustering quality of each individual approach applied on each dataset.
In general, our proposed methods perform reasonably well and show a close match to KPCA K-means and Kernel K-means.
Also, the performance of the proposed approaches is similar to each other.
These factors suggest that the proposed projection approaches possess the JL-Type projection properties.
Furthermore, we find that the proposed approaches in some cases have markedly 
better performance than the Kernel K-means.
One of the possible reasons could be that the random projection  
reduces the eccentricity of original Gaussian-distributed clusters and make clusters in projected spaces more spherical~\cite{Dasgupta00}.
         
Intrinsic K-means gives us reasonable results as it directly works on manifold space.
Compared to the intrinsic approach, LogE has a worse performance in most of datasets.
An exception is on the Ballet dataset where the intrinsic approach has a worse Rand Index than the LogE. 
We conjecture that this is caused by the failure of the intrinsic algorithm to converge in 100 iterations.
Nevertheless, the other performance metrics such as CP, F-Measure and NMI for the intrinsic approach in this dataset still show reasonable performance.
The worse performance for LogE is due to significant distortion of the pairwise distance produced 
when the points are projected into a tangent space.
The G-clustering has a better Rand Index than the intrinsic approach in the Ballet dataset,
which is a similar conclusion drawn in the original work proposing the approach~\cite{shirazietal2012}.
Note that the measurements for clustering performance are different from that in~\cite{shirazietal2012}.
In most cases, the G-clustering is not robust as the performance of G-clustering measured by CP, F-measure and NMI is usually low.
In addition, we do not report the G-clustering results for the UCF101 dataset, as the K-means does not converge within a specified amount of time.
 \begin{figure}
           \centering
                     \includegraphics[width=0.8\columnwidth]{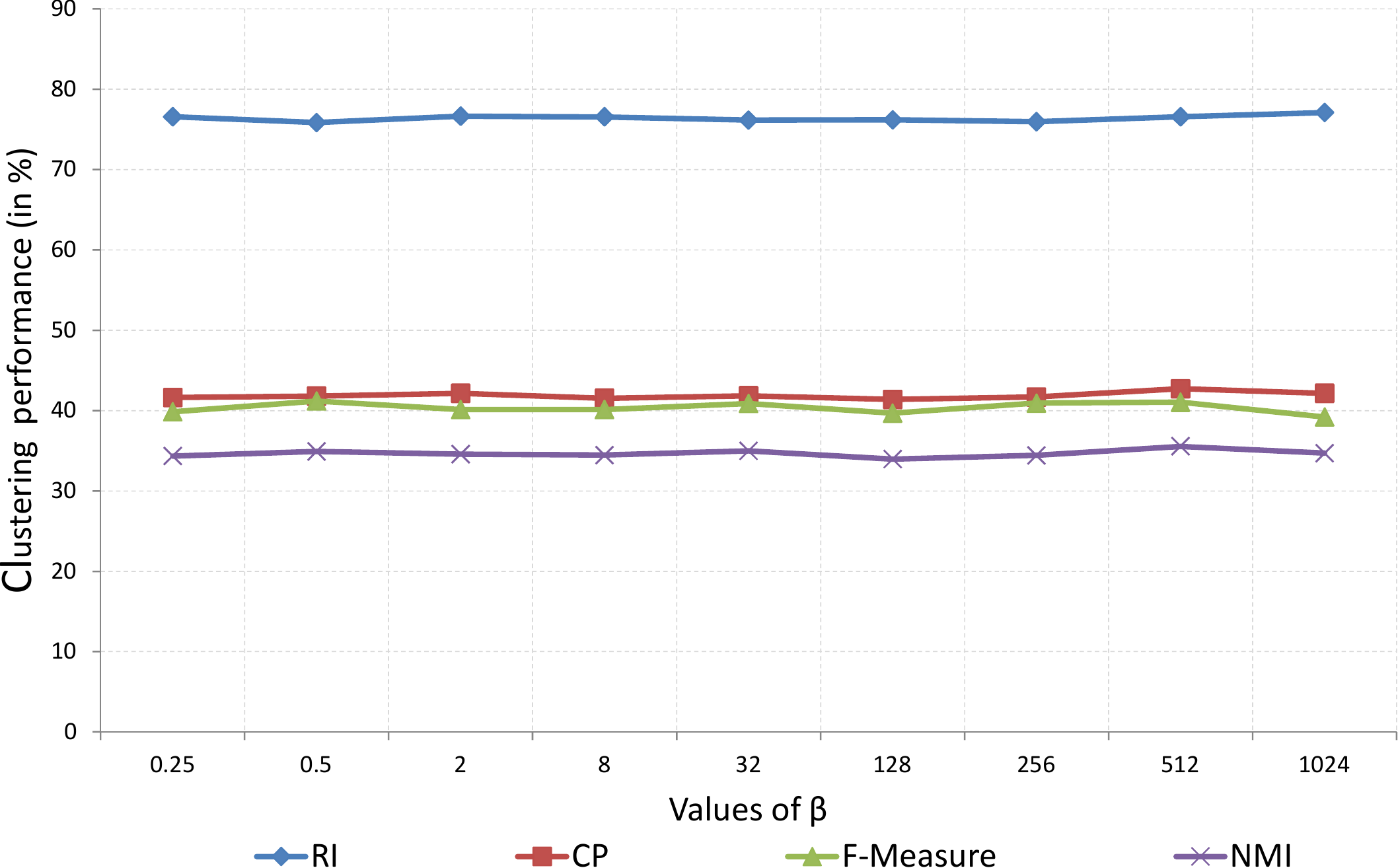}
           \caption
                      { Clustering quality of the proposed KPCA-RP when the kernel parameter $\beta$ was varied on the Ballet dataset. 
                    The clustering quality is measured by:  Rand Index (RI), Cluster Purity (CP), F-Measure and Normalized Mutual Information~(NMI).  }
              \label{fig:p_setting_ballet}         
         \end{figure}
         
\begin{figure}
  \centering
                    \includegraphics[width=0.8\columnwidth]{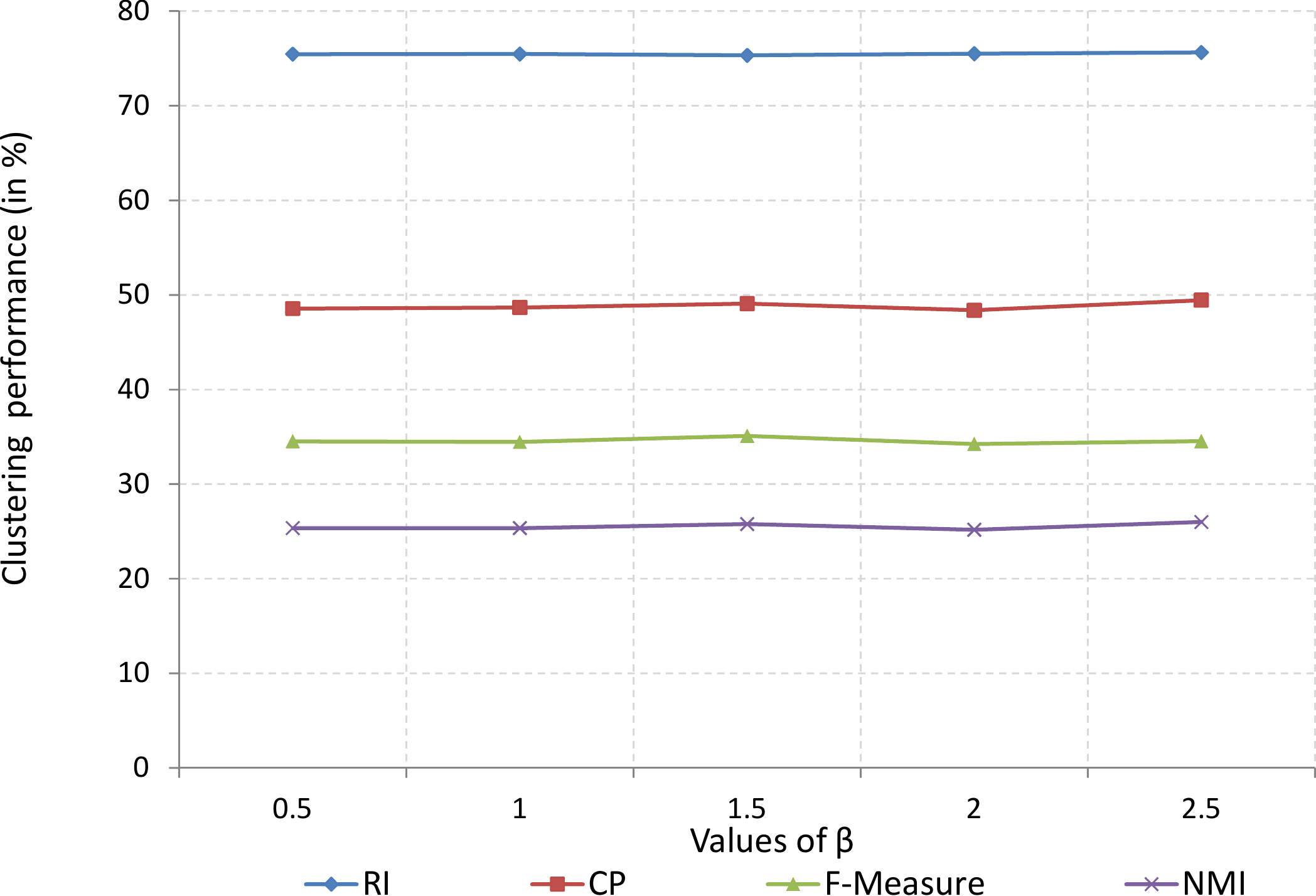}
          \caption{ Clustering quality of the proposed KORP when the parameter $\beta$ is varied on the HEp-2 Cell ICIP2013 dataset.
                               The clustering quality is measured by: Rand Index (RI), Cluster Purity (CP), F-Measure and Normalized Mutual Information~(NMI).} 
                      \label{fig:p_setting_cell}
        \end{figure}

We found the performance of our proposed methods does not change significantly, when the parameters are varied.
Figures~\ref{fig:p_setting_ballet} and~\ref{fig:p_setting_cell} show two examples of the clustering results of KPCA-RP and KORP with different parameters on the  Ballet and HEp-2 Cell ICIP2013 dataset, respectively. We note that the results on the other datasets also exhibit similar trends. This suggests that the issue raised in~\cite{pal2014and}\rev{, where different parameters may adversely alter the kernel space,} may not have significant effect \rev{on} our work. We conjecture that this might be due to the selected manifold kernels crafted to capture the manifold intrinsic structure. 
However, if in the case where the parameter choice of the manifold kernel significantly contributes to the clustering results, one could use a randomly selected small subset of data to perform the parameter search.

The evaluation has clearly shown that our proposal has similar performance to the kernel methods such as KPCA K-means and Kernel K-means.
Indeed, these results alone do not give us much advantage over the other methods.
However, we now present the main advantage of our proposal which is a direct consequence of applying random projection.

\subsection{Run Time Comparative Analysis}

\begin{table}
  \centering
  \caption
    {
    The run time (in seconds) of the approaches on each dataset. Lower run time is better. As in each iteration of K-means, the run time is extremely similar, we report the average run time of each approach without variance. The datasets presented in the first three columns~(\ie Ballet, UCSD and UCF101) are modelled in Grassmannian manifolds, whilst the other three (\ie Brodatz, KTH-TIPS2b and HEp-2 Cell ICIP2013 (shorten as Cell)) are modelled in SPD manifolds. The last three rows are the proposed approaches. SIS and G-clustering are only applicable for SPD manifolds and Grassmannian manifolds, respectively. We refer to Section~\ref{sec:ex_setting} for further explanation of each approach.
    } 
    \label{tab:time}
    \vspace{0.5ex}
\scalebox{0.9}{
    \begin{tabular}{c|ccc|ccc}
    \toprule
        {\bf Methods/Dataset}
          &{\bf Ballet}
          &{\bf UCSD}
          &{\bf {UCF101}}
          &{\bf Brodatz}
          &{\bf KTH-TIPS2b}
          &{\bf {Cell} }  \\
       \toprule
   
    {\bf Intrinsic~\cite{Turagaetal2011}}~&$3966.49$&$1990.02$&${1.64\times 10^5}$&$24.63$&$938.95$&${564.49}$\\ 
       {\bf SIS~\cite{hongetal2009} }&$N/A$&$N/A$&${N/A}$&$4.77$&$60.43$&${185.81}$\\ 
   	 {\bf LogE~\cite{faraki2014fisher} }&$3.35$&$1.55$&${9088.11}$&$\textbf{0.15}$&$\textbf{4.85}$&$\bf{{2.32}}$\\ 
   	 {\bf G-clustering~\cite{shirazietal2012}}&$2.81$&$0.74$&${N/A}$&$N/A$&$N/A$&${N/A}$\\ 
       {\bf Kernel K-means~\cite{dhillon2004kernel,jayasumanaetal2013} }~&$1.06$&$0.70$&${2019.55}$&$22.57$&$675.75$&${2172.87}$\\
       {\bf KPCA~\cite{scholkopfetal1998,jayasumanaetal2013} }~&$1.47$&$0.73$&${6.11\times 10^4}$&$22.42$&$699.34$&${2881.10}$\\ 
       \bottomrule
       {\bf KGRP}&$\textbf{0.51}$&$0.53$&${238.64}$&$7.08$&$14.61$&${21.95}$\\ 
       {\bf KORP}&$0.58$&$\textbf{0.49}$&$\bf{{101.87}}$&$7.03$&$11.75$&${17.73}$\\ 
       {\bf KPCA-RP}&$0.60$ &$\textbf{0.49}$&${102.79}$&$7.75$&$12.28$&${17.73}$\\ 
\bottomrule
\end{tabular}
}
\end{table}

Table~\ref{tab:time} presents the average run time of the individual approach on each dataset.
One of the striking observations from this table is that our proposed approaches have very fast run times.
In some cases (\ie Ballet, UCSD and \rev{UCF101} datasets) they outperform the LogE which is expected to be the fastest method.
The bottleneck suffered by LogE in these datasets is from the high dimensionality of the feature vectors significantly slowing
the K-means algorithm.
Note that, although the run time of LogE on Brodatz, KTH-TIPS2b and  HEp-2 Cell ICIP2013 dataset is quicker than our proposed methods,
the clustering quality shown in Tables~\ref{tab:BRODATZ},~\ref{tab:KTH-TIPS2b} and~\ref{tab:cell} is much worse than that of ours.


The proposed approaches are considerably faster than the kernel approaches such as KPCA K-means and Kernel K-means. 
This is because the proposed approaches only compute the kernel matrix on a small subset of data points.
The benefit will become more pronounced for large datasets such as KTH-TIPS2b, UCF101 and HEp-2 Cell \rev{ICIP2013} datasets where our proposed approach achieves $57.5$ (\ie $\frac{675.75}{11.75} \approx 57.5$), $19.8$ (\ie $\frac{2019.55}{101.87} \approx 19.8$) and $ 122.5$ times (\ie $\frac{2172.87}{17.73} \approx 112.5$) speed up, respectively.
Thus, the proposed approaches will contribute significantly to the clustering of large amount of images or video data for practical applications.

The speed up gained by the proposed approaches is attributed to the effect of 
applying random projection into a reduced projection space.
The proposed approaches also have additional advantages over the kernel approaches as they do not need to compute the kernel matrix on the entire dataset.

In addition, we analyse the computational complexity of each method in Table~\ref{tab:cc}.
In general, each method has two main steps: (1) Data pre-processing and (2) K-means steps. Data pre-processing may include kernel computation and/or projection. Whilst, K-means step comprises cluster membership and cluster mean computations.
In Intrinsic K-means, the pre-processing step is not required. To calculate mean of each cluster, one need to use the intrinsic mean, denoted Karcher mean~\cite{pennec2006intrinsic} that requires multiple iterations to converge. The intrinsic distance is also used for membership computation.  
For LogE, each manifold point needs to be projected onto the Log-Euclidean space. This projection is done once. Then, K-means is applied in the Log-Euclidean space. 
The computational complexity of KPCA and Kernel K-means follows quadratic and cubic growth, respectively. However, our proposed methods have linear growth, as the number of data points, $n$, is much bigger than the size of subset, $p$. This further corroborates the results presented in Table~\ref{tab:time}.

\begin{table}
  \centering
  \caption
    {
   Computational complexity of the approaches on each dataset.
   The dimensionality of SPD and Grassmannian points is $d\times d$ and $q\times d$, respectively. For convenience, $\mathcal{G}$ is used to represent Grassmannian manifold in this table. Note that: $n$ is the number of points; $m$ is the number of clusters; $\ell$ is the number of iterations of K-means; $\ell_{kar}$ is the number of iterations of Karcher mean; $b$ is the dimensionality of the random projection space generated by KGRP \rev{and} $p$ is the dimensionality of the random projection space generated by KORP and KPCA-RP ($p=|\mathcal{S}|$).} 
    \label{tab:cc}
    \vspace{0.5ex}
\scalebox{0.7}{
    \begin{tabular}{|c|ccccc|}
    \toprule {\bf }&{\bf {Compute}}&{\bf{Compute}}&\bf{{Compute}}&{\bf {Compute}}&{\bf {Overall}}\\
      {\bf}&{\bf {Kernel }}&{\bf{Projection}}&\bf{{Mean }}&{\bf {Membership}}&{\bf {Complexity}}\\
     
    \toprule
    {\bf {Intrinsic(SPD)~\cite{Turagaetal2011}}}~&${N/A}$&${N/A}$&${O(\ell \ell_{kar}nd^3)}$&${O(\ell nmd^3)}$&${O(\ell \ell_{kar}nd^3+\ell nmd^3)}$\\ 
    {\bf {Intrinsic($\mathcal{G}$)~\cite{Turagaetal2011}}}~&${N/A}$&${N/A}$&${O(\ell \ell_{kar}n(qd^2+d^3))}$&${O(\ell nm(qd^2+d^3))}$&${O((\ell \ell_{kar}n+\ell nm)(qd^2+d^3))}$\\ 
         
           {\bf{SIS~\cite{hongetal2009} }}&${N/A}$&${O(nd^3)}$&${O(\ell nd^2)}$&${O(\ell nmd^3)}$&${O(\ell nmd^3)}$\\ 
       	 {\bf {LogE(SPD)~\cite{faraki2014fisher}} }&${N/A}$&${O(nd^3)}$&${O(\ell nd^2)}$&${O(\ell nmd^2)}$&${O(nd^3+\ell nmd^2)}$\\ 
       	 {\bf {LogE($\mathcal{G}$)~\cite{faraki2014fisher}} }&${N/A}$&${O(nqd^2)}$&${O(\ell nqd)}$&${O(\ell nmqd)}$&${O(nqd^2+\ell nmqd)}$\\ 
       	 {\bf {G-clustering~\cite{shirazietal2012}}}&${O(n^2)}$&${O(n^3)}$&${O(\ell n^2)}$&${O(\ell n^2m)}$&${O(n^3+\ell n^2m)}$\\ 
           {\bf{ Kernel K-means~\cite{dhillon2004kernel,jayasumanaetal2013}} }~&${O(n^2)}$&${N/A}$&${N/A}$&${O(\ell n^2m)}$&${O(\ell n^2m)}$\\
           {\bf {KPCA~\cite{scholkopfetal1998,jayasumanaetal2013}} }&${O(n^2)}$&${O(n^3)}$&${O(\ell n^2)}$&${O(\ell n^2m)}$&${O(n^3+\ell n^2m)}$\\ 
           \bottomrule
           {\bf{ KGRP}}&${O(np)}$&${O(p^3+np^2)}$&${O(\ell npb)}$&${O(\ell nmb)}$&${O(np+p^3+np^2+\ell nmb)}$\\ 
           {\bf{ KORP}}&${O(np)}$&${O(p^3+np^2)}$&${O(\ell np)}$&${O(\ell nmp)}$&${O(np+p^3+np^2+\ell nmp)}$\\ 
           {\bf {KPCA-RP}}&${O(np)}$&${O(p^3+np^2)}$&${O(\ell np)}$&${O(\ell nmp)}$&${O(np+p^3+np^2+\ell nmp)}$\\ 
\bottomrule
\end{tabular}
}
\end{table}
\subsection{Further Analysis}

In this section, we analyse the parameters contributing to the performance and run time of the proposed methods.
\rev{Due to space limitations, we only show the performance measured by RI and CP. Note that the performance measured by F-Measure and NMI also follows the same trends.} An obvious parameter is the projected space dimensionality, $k$.
When $k$ is small, each data point will be represented in a much smaller feature vector, resulting in faster K-means clustering processes.
Another parameter is $|\mathcal{S}|$, the size of set $\mathcal{S}$ which determines the run time of the kernel matrix computation.
As $|\mathcal{S}|$ gets larger, it takes longer to compute the kernel matrix.
Smaller $|\mathcal{S}|$ gives more advantage to the proposed methods over the kernel approaches such as Kernel K-means and KPCA that require kernel computation on the entire data points.
We note that $k$ and $|\mathcal{S}|$ have an interesting relationship.
More precisely, for KORP and KPCA-RP, $|\mathcal{S}|$ determines the projected space dimensionality, $k$.
Therefore, it is desirable to make $|\mathcal{S}|$ as small as possible whilst still preserving as much of the pairwise distance.
\begin{figure}[!t]
  \centering
  \includegraphics[width=0.8\columnwidth]{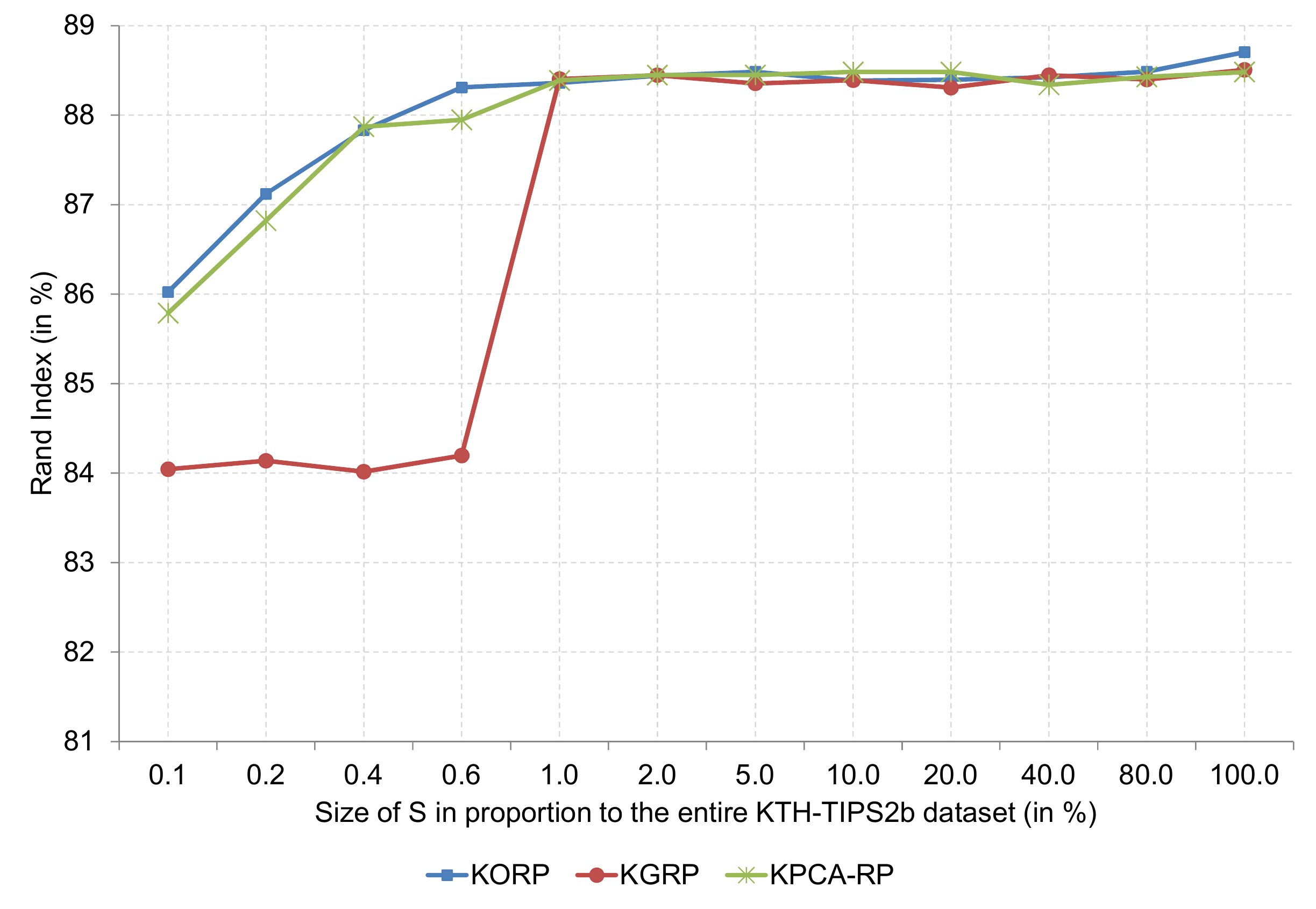}
 \caption
   {
    The Rand Index (in \%) of the proposed approaches when the size of set $\mathcal{S}$ is progressively increased on the KTH-TIPS2b dataset. KGRP: Kernelised Gaussian Random Projection; KORP: Kernelised Orthonormal Random Projection; KPCA-RP: Kernel PCA based Random Projection.
    }
  \label{fig:KTH_RI}
\end{figure} 

 \begin{figure}[!t]
  \centering
  \includegraphics[width=0.8\columnwidth]{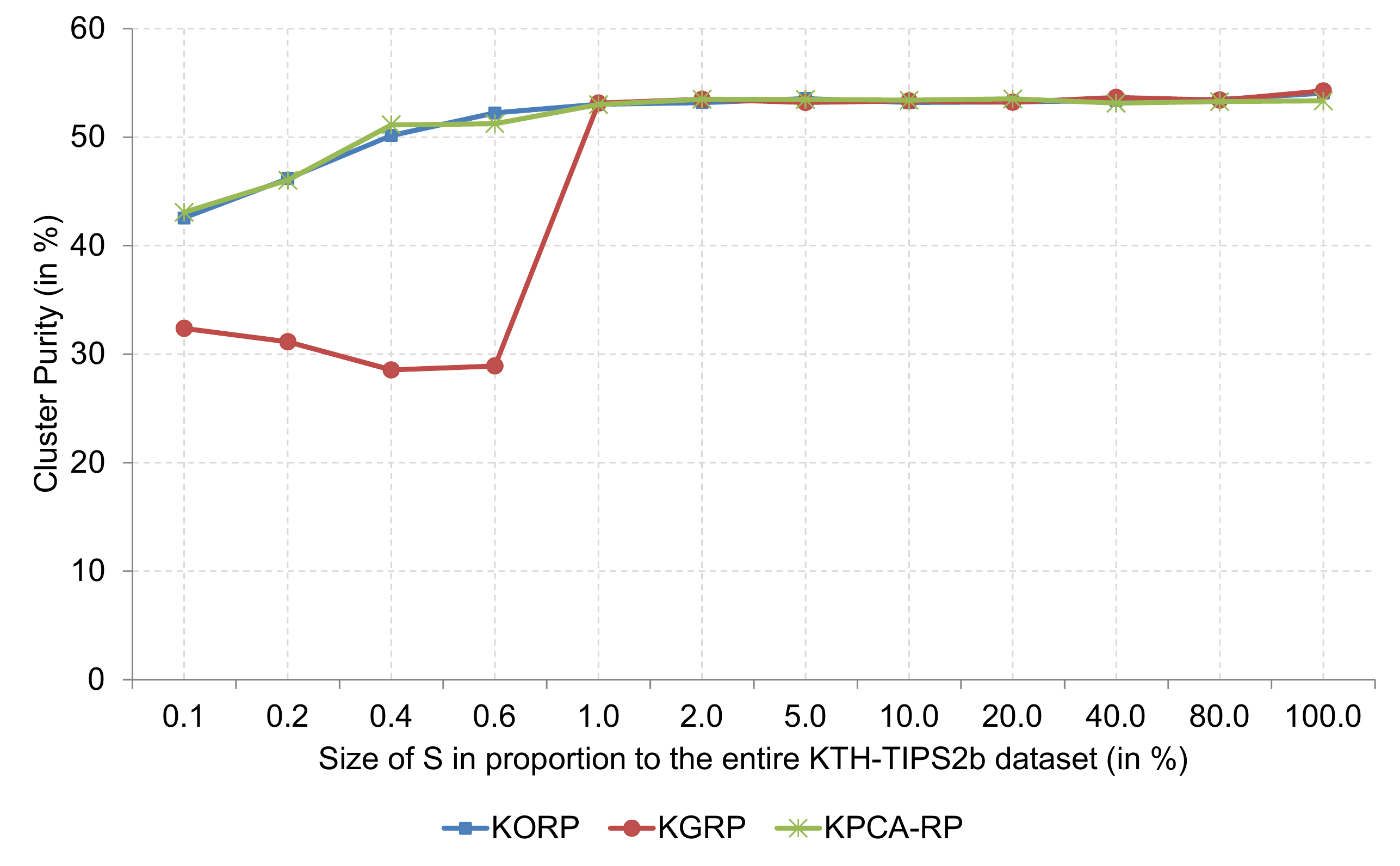}
 \caption
   {
    The Cluster Purity (in \%) of the proposed approaches when the size of set $\mathcal{S}$ is progressively increased on the KTH-TIPS2b dataset. KGRP: Kernelised Gaussian Random Projection; KORP: Kernelised Orthonormal Random Projection; KPCA-RP: Kernel PCA based Random Projection.
    }
  \label{fig:KTH_CP}
\end{figure}

In contrast to KORP and KPCA-RP, KGRP separates the projected space dimensionality to $|\mathcal{S}|$.
Nevertheless, we found that $|\mathcal{S}|$ still plays an important role in the overall system performance.
To verify this, we vary $|\mathcal{S}|$ on the KTH-TIPS2b.
As we can see from Figures~\ref{fig:KTH_RI} and~\ref{fig:KTH_CP}, the performance of the proposed approaches  
increases as $|\mathcal{S}|$ is progressively increased.
The performance increase stops when $|\mathcal{S}|$ reaches a particular value.
In this analysis we also found that the performance of KORP and KPCA-RP is 
markedly better than KGRP when $|\mathcal{S}|$ is considerably small.
A possible reason is that the CLT requires the set $\mathcal{S}$ to have a minimum number of elements~(normally 30) in order
to make the theorem applicable.

The above observation suggests the following facts about $|\mathcal{S}|$: (1)~$|\mathcal{S}|$ determines the run time for all the 
proposed approaches (\ie on the kernel computation); (2)~$|\mathcal{S}|$ also contributes to the K-means run time for KORP and KPCA-RP;
(3)~the lower bound of $|\mathcal{S}|$ in the KGRP is related to the 
lower bound of the CLT and (4) the lower bound of $|\mathcal{S}|$ for KORP and KPCA-RP is related to the lower bound of $k$.

\begin{figure}
  \centering
  \includegraphics[width=0.8\columnwidth]{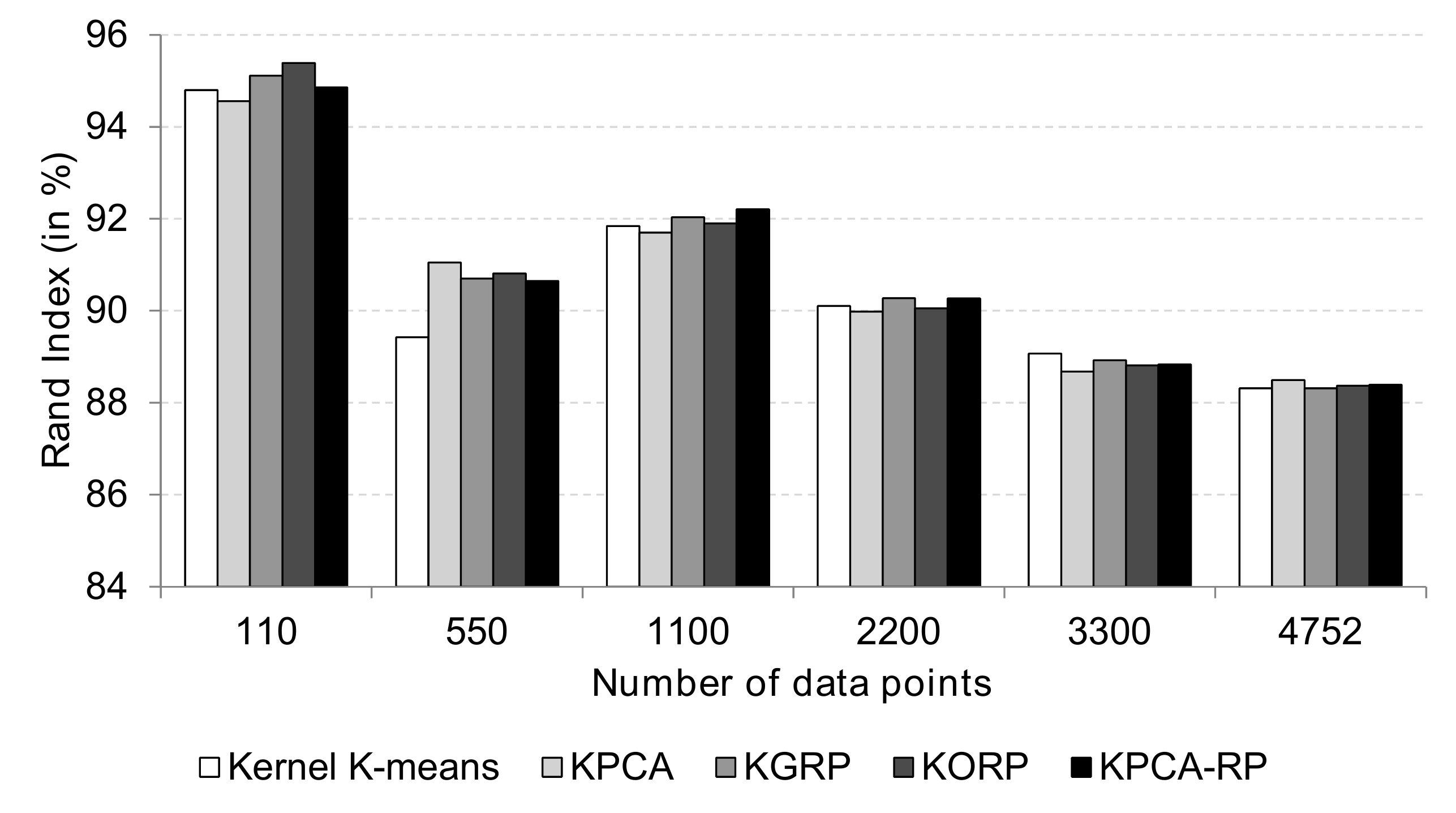}
 \caption
   {
		The Rand Index (in \%) of the proposed approaches, Kernel K-Means and KPCA applied on subsets of KTH-TIPS2b with various sizes. 
		We fix $|\mathcal{S}|$ for all subsets.
    }
  \label{fig:KTH_RI_2}
\end{figure}

\begin{figure}[!t]
  \centering
  \includegraphics[width=0.8\columnwidth]{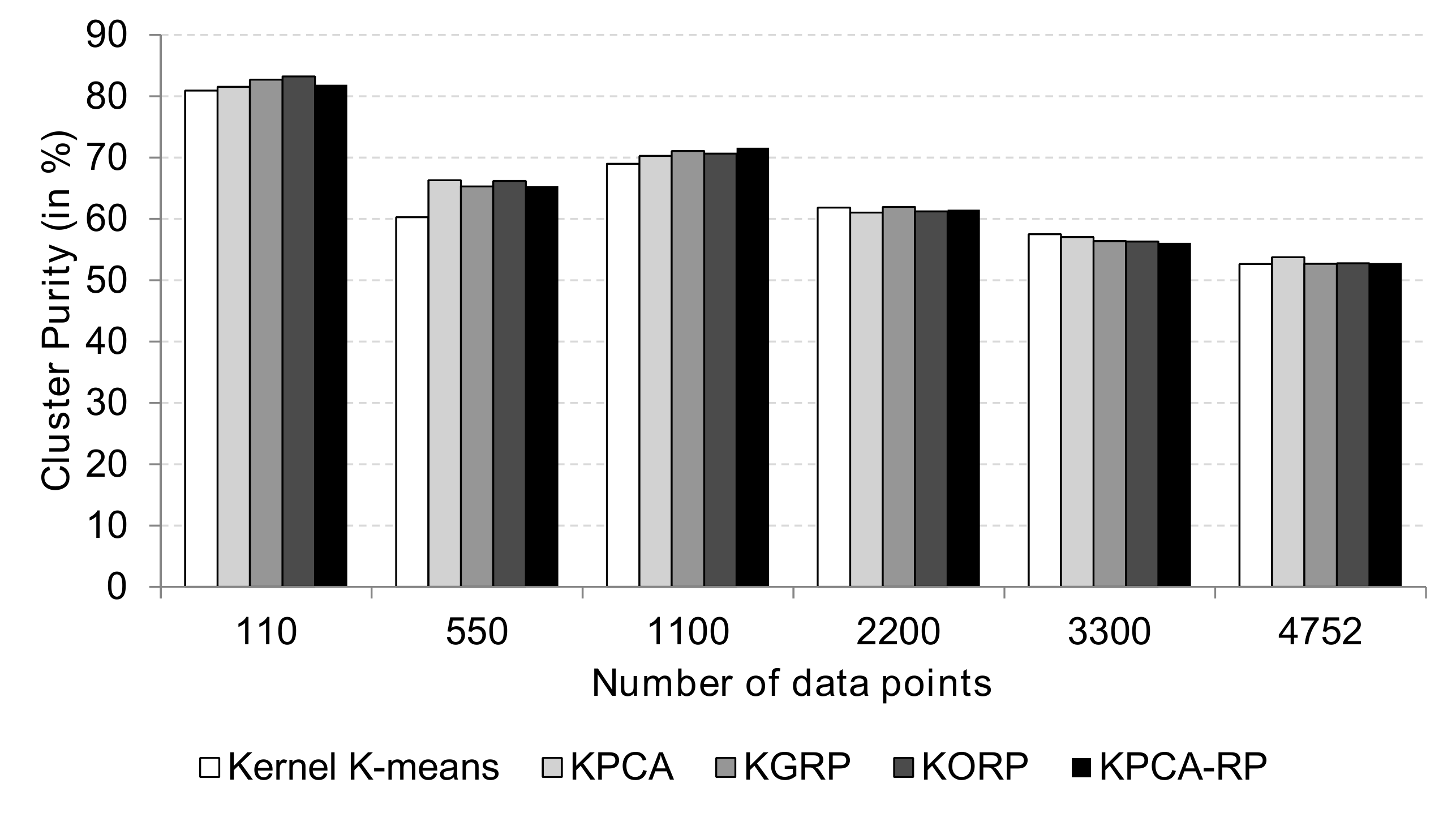}
 \caption
   {
		The Cluster Purity (in \%) of the proposed approaches, Kernel K-Means and KPCA applied on subsets of KTH-TIPS2b with various sizes. 
		We fix $|\mathcal{S}|$ for all subsets.
    }
  \label{fig:KTH_CP_2}
\end{figure}

The JL-Lemma relates $k$ to the total number of data points, $n$ (refer to Lemma~\ref{JL}). 
This relationship seems unfavourable for KORP and KPCA-RP as this would mean $|\mathcal{S}|$ increases as $n$ increases.
Fortunately, Lemma~\ref{blum} and Theorem~\ref{theorem:kpcarp} suggest that $k$ is related to the margin between classes.
This means that we now need only consider the separating margin to select $|\mathcal{S}|$.
To further corroborate this empirically, we apply the proposed approaches by varying the dataset size of the KTH-TIPS2b.
We assume that the margin is relatively unchanged though the dataset size is varied.
More precisely, we first fix $|\mathcal{S}|$ for each proposed approach.
Then we randomly select the data points from the KTH-TIPS2b to create a smaller version of the dataset.
The proposed approaches are applied on these smaller subsets of the dataset.
Note that although $|\mathcal{S}|$ is fixed, we still select the elements of $\mathcal{S}$ from the given subset.
The results shown in Figures~\ref{fig:KTH_RI_2} and~\ref{fig:KTH_CP_2} suggest that the proposed approaches still have on par performance with both Kernel K-means and KPCA K-means,
suggesting that $|\mathcal{S}|$ relates to the margin separation between classes.

%% file: conclusion.tex
\section{Conclusions and Future Directions}
\label{sec:conclusions}
Clustering over Riemannian manifolds plays an important role in the automatic analysis of images and videos~\cite{Turagaetal2011,shirazietal2012}.
As discussed before, in general, the existing methods suffer from either poor performance or high computational complexity.
In this paper, we propose a novel framework with random projection to tackle the clustering problems over Riemannian manifolds.
Based on the framework, we present three random projection methods for manifold points:~KGRP, KORP and KPCA-RP.
Through experiments on several computer vision applications, we demonstrate that our proposed framework achieves significant speed increases while maintaining clustering performance in comparison to the other conventional methods such Kernel K-means.
Furthermore, we analyse the parameters that impact the performance and run time of our proposed methods.


In the proposed framework, we carry out random projection for manifold points with the aid of RKHS.
In other words, we first project manifold points into RKHS. 
One promising future direction is to study the intrinsic random projection, which directly maps manifold points into the random projection space.
To this end, one needs to define the notions of projection and hyperplane generation process in the manifold space.

%% file: Acknowledgements.tex
\\
\\
\noindent
\textbf{Acknowledgements}

The authors thank Danny Smith for his helpful suggestions and comments to improve the paper.
They also thank the anonymous reviewers for their insight and guidance on improving this manuscript.
This research was partly funded by Sullivan Nicolaides Pathology, Australia and the Australian Research Council (ARC) Linkage Projects Grant LP130100230.

%% file: cv.tex
\vspace{8pt}
{
\textbf{Kun Zhao} received her MSc from University of Electronic Science and Technology of China in 2013.
Currently, she is a PhD student in The University of Queensland (UQ).
Her research interests are in the areas of computer vision, machine learning and pattern recognition.

\textbf{Azadeh Alavi} currently is a research fellow at University of Maryland. She received her PhD from UQ in 2014. She obtained her Bachelor of Applied Mathematics degree in 2002 and worked in industries for about 2 years before commencing her Master of IT-advanced program (Research Stream) at Griffith University. Currently, she is a research fellow at University of Maryland.
Her interests are in the areas of machine learning, pattern recognition and image processing.

\textbf{Arnold Wiliem} is a research fellow at UQ. He received his PhD in 2010 from Queensland University of Technology. He is a member of the IEEE and served as reviewer in various computer vision venues. His current research interests are in the areas of statistical methods in non-linear manifolds, machine learning and pattern recognition.

\textbf{Brian Lovell} received his PhD in 1991 from UQ. Professor Lovell is Director of the Advanced Surveillance Group at UQ. He was President of the International Association for Pattern Recognition (IAPR) [2008-2010], and is Fellow of the IAPR, Senior Member of the IEEE.
His interests include Biometrics, Nonlinear Manifold Learning, and Pattern Recognition.}